\PassOptionsToPackage{table}{xcolor}
\documentclass{article} \usepackage{iclr2027_conference,times}

\usepackage{amsmath,amsfonts,bm}

\def\eqref#1{equation~\ref{#1}}

\def\1{\bm{1}}

\DeclareMathAlphabet{\mathsfit}{\encodingdefault}{\sfdefault}{m}{sl}
\SetMathAlphabet{\mathsfit}{bold}{\encodingdefault}{\sfdefault}{bx}{n}

\input{multilingual_setup}
\usepackage{etoolbox}
\makeatletter
\patchcmd{\@maketitle}{{\LARGE\sc \@title\par}}{{\LARGE\sc\fontencoding{T1}\fontfamily{ptm}\selectfont\raggedright\hyphenpenalty=10000 \exhyphenpenalty=10000 \@title\par}}{}{\PackageWarning{paper}{title font patch did not apply}}
\makeatother

\usepackage{hyperref}
\usepackage{url}
\usepackage{graphicx}
\graphicspath{{figs/}}
\usepackage{wrapfig}
\usepackage{booktabs}
\usepackage{tabularx}
\usepackage{threeparttable}
\usepackage{multirow}
\usepackage{xcolor}
\usepackage{adjustbox}
\usepackage{ragged2e}
\usepackage{array}
\usepackage{placeins}
\usepackage{makecell}
\usepackage{xspace}
\usepackage{algorithm}
\usepackage{algpseudocode}
\newcolumntype{Y}{>{\RaggedRight\arraybackslash}X}
\usepackage[tableposition=top]{caption}

\title{Linguistic Loopholes in LLM Unlearning: From a 174-Language Benchmark to Coverage-Aware Unlearning}

\author{Tyler Skow, Shravan Chaudhari, Rama Chellappa \& Abhay Yadav \\
Johns Hopkins University \\
\href{mailto:tskow1@jh.edu}{\texttt{tskow1@jh.edu}}
}

\newcommand{\Lall}{\mathcal{L}}
\newcommand{\Lsrc}{\mathcal{L}_{\mathrm{src}}}
\newcommand{\Lheld}{\mathcal{L}_{\mathrm{held}}}
\newcommand{\Fset}{\mathcal{F}}
\newcommand{\Rset}{\mathcal{R}}

\newcommand{\benchmark}{\textsc{Cross-lingual Unlearning Tensor}\xspace}
\newcommand{\method}{\textsc{COVER}\xspace}

\newcommand{\gptj}{GPT-5.4\xspace}
\newcommand{\qwenj}{Qwen3.5-27B\xspace}

\newcommand{\sdv}[1]{{\scriptsize$\pm$#1}}

\makeatletter
\newcommand{\inputhere}[1]{\begingroup
  \let\inputhere@float\@float
  \def\@float##1{\@ifnextchar[{\inputhere@skip{##1}}{\inputhere@float{##1}[H]}}\def\inputhere@skip##1[##2]{\inputhere@float{##1}[H]}\input{#1}\endgroup}
\makeatother

\newcommand{\apppart}[2]{\addlinespace[5pt]\multicolumn{4}{@{}l}{\textsc{Part #1}\quad\textbf{#2}}\\\addlinespace[2pt]}
\newcommand{\appsec}[3]{\ref*{#1} & \nameref{#1}\if\relax\detokenize{#3}\relax\else\newline{\footnotesize\color{black!65}#3}\fi & #2 & \pageref{#1}\\\addlinespace[3pt]}
\newcommand{\appsub}[1]{\mbox{\ref*{#1}~\nameref{#1}}}
\newcommand{\appsep}{\nobreak\hspace{.5em}\textperiodcentered\hspace{.5em}\ignorespaces}

\iclrfinalcopy
\begin{document}

\maketitle
\vspace{-\baselineskip}

\begin{abstract}
Unlearning a fact in one language does not guarantee its removal in others as changing the query or even the requested answer language can reopen seemingly forgotten knowledge—a cross-lingual loophole. The most straightforward solution to this challenge -- unlearning in all languages -- is neither scalable nor desirable as it amplifies damage to unrelated model capabilities. We introduce the task of language budgeted multilingual unlearning where the goal is to select a subset of languages that maximizes cross-lingual erasure. To study this task we introduce \benchmark, an unlearning benchmark that spans 174 language--script pairs and 25 atomic paraphrase types to examine when forgetting generalizes across linguistic expressions of the same knowledge. We further propose \method, which selects source languages to maximize predicted COVERage of languages receiving no forget supervision, enabling unlearning on a language budget. Surprisingly, we find naively selecting strong individual sources does not reliably compose into strong source sets motivating our development of \method. At deployment \method only requires benign calibration data and access to the frozen model. Across three model families and two disjoint forget sets, \method reduces mean held-out residual access by 7.8--27.3\% relative to uniform source selection. We find these gains extend beyond synthetic benchmarks to real news documents in low-resource language settings using human translated data from the Low Resource Languages for Emergent Incidents (LORELEI) corpus. \footnote{The full dataset will be released after peer review.\\See \href{https://tskow99.github.io/crosslingual-unlearning-tensor/}{data subset} and \href{https://github.com/tskow99/crosslingual-unlearning-tensor}{code}.}

\end{abstract}
\section{Introduction}
\begin{wrapfigure}[18]{r}{0.46\textwidth}
\vspace{-12pt}
\centering
\includegraphics[width=0.46\textwidth,height=157.2306pt,keepaspectratio]{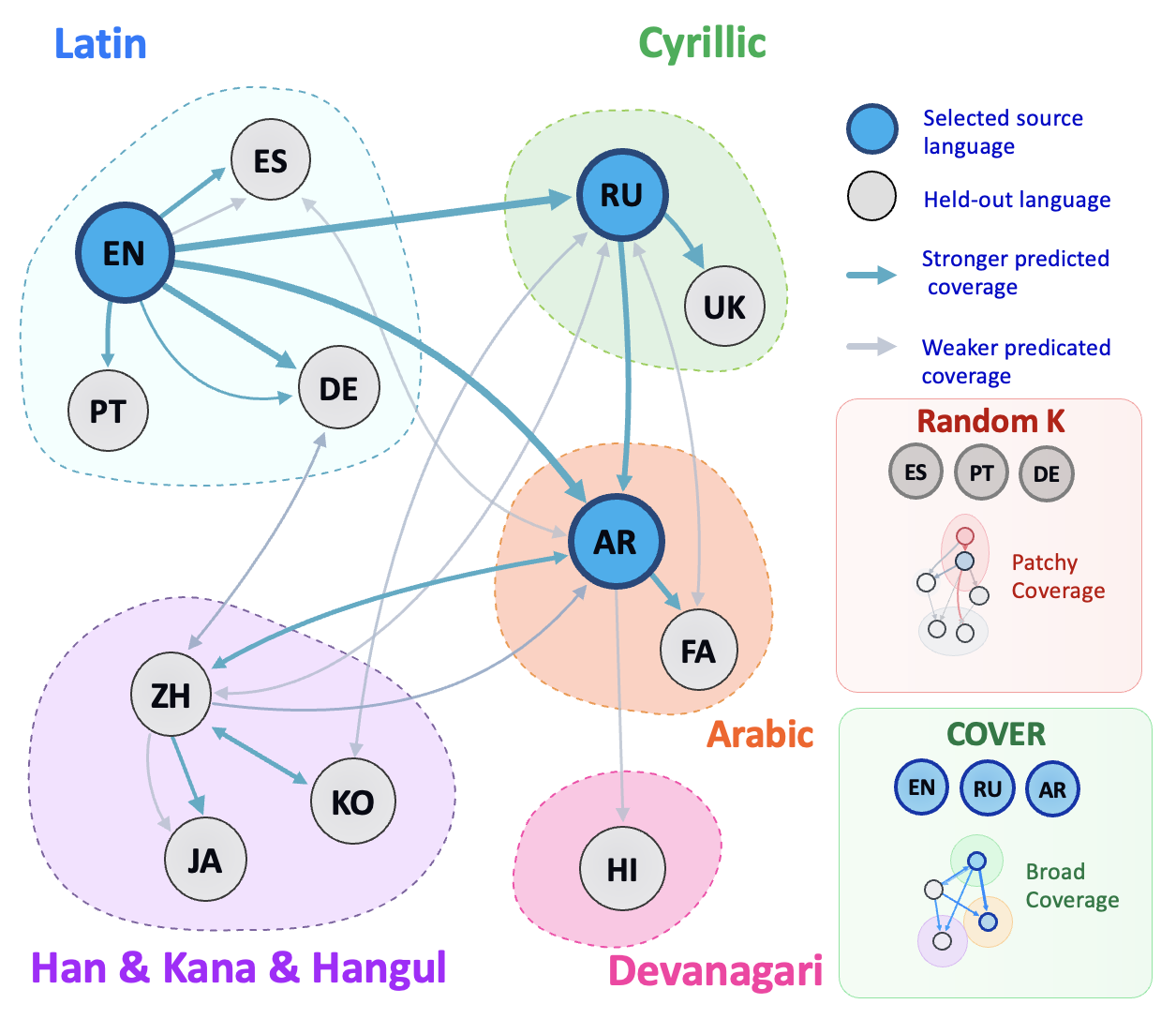}
\caption{\textbf{Language-budgeted unlearning.} \method selects $K$ source languages (blue) to cover held-out languages (grey). Shaded regions group languages by script for visual orientation. Connection and spatial layout are schematic.}
\label{fig:teaser}
\end{wrapfigure}

Large language models can memorize and reproduce data seen during pretraining and fine tuning, including private content, copyrighted material, or knowledge that later becomes unsafe \citep{carlini2021extracting,lukas2023analyzing,lee2023language,chang2023speak,li2024wmdp}. Guardrails and inference time interventions can mitigate unwanted behavior, but they provide limited protection when users have white-box access to model parameters. Retraining from scratch without the offending data offers the strongest possible remedy, but it's typically impractical given the resources required. This motivates LLM unlearning: update a model trained on data \(\mathcal{D}\) so that its behavior on a forget set \(\mathcal{F}\subseteq\mathcal{D}\) approximates a model trained without \(\mathcal{F}\) while preserving utility on retained data \(\mathcal{R}\).

As frontier language models improve in multilingual generation and understanding, so too does the surface area for accessing memorized knowledge across languages \citep{zhao-etal-2024-tracing,qi-etal-2023-cross}. This creates a potential challenge for Machine Unlearning (MU), as removal in one language may not guarantee forgetting in another. Recent works have documented this cross-lingual failure mode \citep{farashah-etal-2026-multilingual, xiang2026multilingual, choi-etal-2024-cross,lu-koehn-2025-learn}, but existing studies cover a limited set of languages which tend to be high resource.

We release \benchmark, a multilingual unlearning dataset spanning 174 language--script pairs. Through the release of \benchmark, we study the dynamics of language and unlearning from five perspectives: the language the fact is learned in, the language(s) used to unlearn it, the language of the evaluation query, the language of the scored answer and how various same language paraphrases differ in terms of eliciting unlearned knowledge.

\benchmark shows that unlearning in one language can leave the same facts accessible through other query--answer routes. Full parallel forget data are often unavailable, and translating each request into every evaluation language increases cost. In our experiments, all-language supervision also increases damage to retained facts and multilingual generation (\autoref{tab:all-language-supervision-budget}).

We therefore study \textit{language budgeted multilingual unlearning}. Forget supervision is available only in a source subset \(\Lsrc\subset\Lall\), with \(|\Lsrc|=K\). We evaluate forgetting in held-out languages \(\Lheld=\Lall\setminus\Lsrc\), alongside retained knowledge and multilingual capabilities.

We introduce \method~(COVerage-guided source selection for multilingual ERasure), which uses only benign calibration data to select sources for expected held-out coverage before applying an unlearning objective. It requires no held-out forget examples at deployment. \method improves held-out forgetting and can reduce generated disclosure more than selecting the three strongest individual sources. Furthermore, we validate the performance of \method on non-synthetic data and human generated translations through the Low Resource Languages for Emergent Incidents (LORELEI) corpus of news documents \citep{strassel-tracey-2016-lorelei}. On Aya \citep{salamanca2026tinyayabridgingscale} and Qwen \citep{yang2025qwen3technicalreport} model families, source selection developed on \benchmark improves cross-lingual forgetting on our constructed LORELEI document removal task spanning eight languages. We summarize our contributions as follows:

\begin{enumerate}
    \item Constructed \benchmark, a dataset with 174 language--script pairs (diverse scripts and resource levels), 25 atomic paraphrase types, translated paraphrases and evaluation on query and answer routes separately, enabling a systematic study of how language and wording affect unlearning.
    \item Formalized the language-budgeted multilingual unlearning setting where we need to select \(K\) source languages for forgetting that generalizes to held-out languages while preserving retained knowledge and multilingual capabilities.
    \item We show that meeting the removal criterion on a supervised route does not guarantee multilingual forgetting and characterize residual access across query and answer language routes.
    \item We propose \method, which selects sources without held-out forget examples at deployment and reduces held-out residual access across existing benchmarks and real-world low-resource settings.
\end{enumerate}

\label{sec:introduction}
\section{Problem formulation and evaluation protocol}
\label{sec:protocol}

\textbf{Unlearning}
Let $M$ be an LM trained on distribution $D$. In unlearning, we are given a \textit{forget set} \(\Fset\) $\subset D$ and a \textit{retain set} \(\Rset\subset D\), with the goal of producing an unlearned model $M'$ that behaves as if it had been trained on $D \setminus \Fset$. We use \(\Rset\) to measure drift from capabilities we wish the model to retain  \citep{unlearning:sp15}.
\\
\textbf{Multilingual Unlearning~~}
Let \(x=(q_{\ell},a_{\ell},s)\) denote a sample with query language \(q_{\ell}\), target answer language \(a_{\ell}\), and unlearning target \(s\), where \(s\) groups paraphrases that query the same underlying fact. Under a language budgeted unlearning request, forget examples are available only in source languages \(\Lsrc\subset\Lall\), where $\Lall$ is a set of all languages $\ell$, \(|\Lsrc|=K\), while evaluation covers both \(\Lsrc\) and held out languages \(\Lheld=\Lall\setminus\Lsrc\). A \emph{route} $\ell_q\to\ell_a$ specifies the query and answer languages; a \emph{native route} uses the same language for both.
\paragraph{Notation}
For an example $x$ let $p_{\text{M}}(x)$ be the teacher forced answer likelihood of the gold answer. For an evaluation route, $p_{\text{M}}$ denotes the mean of these scores. $\mathrm{FT}$, $\mathrm{O}$ and $\mathrm{U}$ denote the finetuned, retain only oracle and unlearned models. We define oracle normalized residual access as \(R=\frac{p_{\mathrm{U}}-p_{\mathrm{O}}}{p_{\mathrm{FT}}-p_{\mathrm{O}}},\) and oracle normalized removal $= 1-R$. $R=0$ matches oracle, $R=1$ matches finetuned model. $100R$ expresses remaining likelihood gap above retain oracle as a percentage of original vs. finetuned-oracle gap.

\paragraph{Evaluation protocol}
We use a matched source forget protocol so methods are compared after reaching the same amount of source-language forgetting. Without this control, a method can look effective because it overforgets and produces gibberish output, or appear to preserve retained knowledge because it has not forgotten. We evaluate checkpoints at dense intervals up to a fixed number of optimization steps and select the first checkpoint reaching an oracle normalized progress threshold, inspired by \(MU95\) in \citep{lee-etal-2025-localization}. Source progress measures the fraction of the fine-tuned vs. oracle likelihood gap removed at a checkpoint. Let \(S=\Lsrc\), $t$ index a checkpoint, and $p_{t,\ell}$ denote mean forget answer likelihood on the source route $\ell \rightarrow \ell$. Define $\operatorname{Progress}_{t,\ell}=(p_{\mathrm{FT},\ell}-p_{t,\ell})/(p_{\mathrm{FT},\ell}-p_{\mathrm{O},\ell}+\epsilon)$. We select the first checkpoint satisfying
\begin{equation}
\frac{1}{|S|}\sum_{\ell\in S}\operatorname{Progress}_{t,\ell}\geq0.95,
\qquad \min_{\ell\in S}\operatorname{Progress}_{t,\ell}\geq0.90.
\label{eq:progress-criteria}
\end{equation}

Since forgetting is matched on source language the evaluation is mean and maximum oracle normalized residual access $R$ across held-out routes. We also evaluate on extraction strength \citep{dorna2025openunlearningacceleratingllmunlearning} and finally open ended generation, which we classify as full leakage, partial leakage, code-switching and gibberish output via an LLM judge. We also include chrF++ scores to measure text similarity \citep{popovic2017chrf++}. \autoref{app:generation-judge} details the generation judge and the extraction-strength score.

\section{\benchmark}
\label{sec:benchmark}
\begin{table}[!htbp]
\centering
\caption{\textbf{The \benchmark at a glance.} Coverage across language, expression, and model axes; scored routes are defined in the text.}
\label{tab:benchmark-at-a-glance}
\small
\setlength{\tabcolsep}{5pt}
\renewcommand{\arraystretch}{1.1}
\begin{tabularx}{\textwidth}{@{}>{\raggedright\arraybackslash}p{0.21\textwidth}
  >{\centering\arraybackslash}p{0.09\textwidth} >{\raggedright\arraybackslash}X@{}}
\toprule
\textbf{Axis} & \textbf{Size} & \textbf{Composition} \\
\midrule
Language--script pairs & 174 & 18 scripts; 27 high-, 46 medium-, 101 low-resource; scored in both roles \\
Question forms & 25 & ETPC atomic paraphrase types in seven meta-categories
  \citep{kovatchev-etal-2018-etpc}; seven types translated into 40 languages
  (1{,}690 rows per language) \\
Fine-tuned checkpoints & 103 & 81 monolingual (27 languages per model family) and 22 multilingual
  configurations over Tiny Aya \citep{salamanca2026tinyayabridgingscale},
  Qwen3 \citep{yang2025qwen3technicalreport} and Llama\,3
  \citep{grattafiori2024llama3herdmodels}, each with a matched retain-only oracle \\
Unlearned checkpoints & 287 & 243 monolingual endpoints ($81\times3$ methods:
  SimNPO \citep{fan2025simplicity}, RMU \citep{li2024wmdp}, GradDiff \citep{dorna2025openunlearningacceleratingllmunlearning})
  and 44 Aya trilingual endpoints, stopped at matched source forgetting \\
\bottomrule
\end{tabularx}
\end{table}

Our benchmark enables the study of multilingual unlearning with 174 language--script pairs across resource tiers. Moreover, we release forget samples re-expressed through 25 syntactically diverse paraphrase types and then translations of those paraphrases (translated paraphrases) into 40 of these languages. We use TOFU \citep{maini2024tofutaskfictitiousunlearning} and LUME \citep{ramakrishna-etal-2025-lume} as the building blocks for our dataset.
Task of Fictitious Unlearning (\textbf{TOFU})  \citet{maini2024tofutaskfictitiousunlearning} is synthetic author-profile QA with 1\%, 5\%, and 10\% forget splits and matching retain splits; we use 10\%. LLM Unlearning with Multitask Evaluations (\textbf{LUME}) \citet{ramakrishna-etal-2025-lume} contains creative short novels, biographies with PII and real biographies from Wikipedia. We finetune on all of these tasks.

For external validation on natural text, we construct, to the best of our knowledge, the first unlearning benchmark on the Low Resource Languages for Emergent Incidents program \citep{strassel-tracey-2016-lorelei}. The subset of the LORELEI corpus we use consists of news web pages. Our LORELEI benchmark consists of 100 documents with human translations in English, Hindi, Spanish, Russian, Chinese, Swahili, Bengali, and Vietnamese. The translations provide matched content across languages, allowing us to study source-language choice beyond synthetic QA.

An entry in \benchmark is a forget fact, learned, unlearned, queried, and answered under different chosen languages, paraphrases and translated paraphrases. We colloquially refer to a given combination of query and answer representation as a \textit{route}. The behavior of a route varies depending on languages used for learning and unlearning. Each cell stores $p_M(x)$, the length normalized teacher forced probability of the gold answer for the sample $x=(q_{\ell_q},a_{\ell_a},s)$ of \autoref{sec:protocol}.

We sparsely score five query--answer grids through acquisition language $a$ and English: $\ell\to a$ and $a\to\ell$ vary the query and answer language, respectively, over 174 language--script pairs; $\ell\to\mathrm{en}$ and $\mathrm{en}\to\ell$ repeat these controls through English; and $\ell\to\ell$ tests native-language access in the 173 other language--script pairs.

In \autoref{tab:benchmark-at-a-glance} we provide a breakdown of our full benchmark coverage and \autoref{tab:ept_paraphrase_types} summarizes our paraphrase taxonomy. The full language set and complete paraphrase type definitions are in \autoref{app:benchmark}. We finetune each model family (Tiny Aya \citep{salamanca2026tinyayabridgingscale}, Qwen3 \citep{yang2025qwen3technicalreport} and Llama\,3 \citep{grattafiori2024llama3herdmodels}) in 27 languages, generating 81 monolingual checkpoints. The 22 multilingual checkpoints comprise seven three-language mixtures for Tiny Aya and five shared five-/six-language mixtures per model family, bringing the total to 103 fine-tuned checkpoints, each with a matched retain-only oracle. Unlearning produces 287 checkpoints: 243 from the monolingual checkpoints (three methods per checkpoint) and 44 from the Aya trilingual checkpoints. \autoref{tab:model-checkpoint-coverage} contains the details on these language sets. For the details on how we generated our translations and paraphrases, please refer to \autoref{app:translation} and \autoref{app:paraphrase}.

\subsection{Findings from the \benchmark}
\label{sec:findings}
\begin{figure}[t]
\centering
\includegraphics{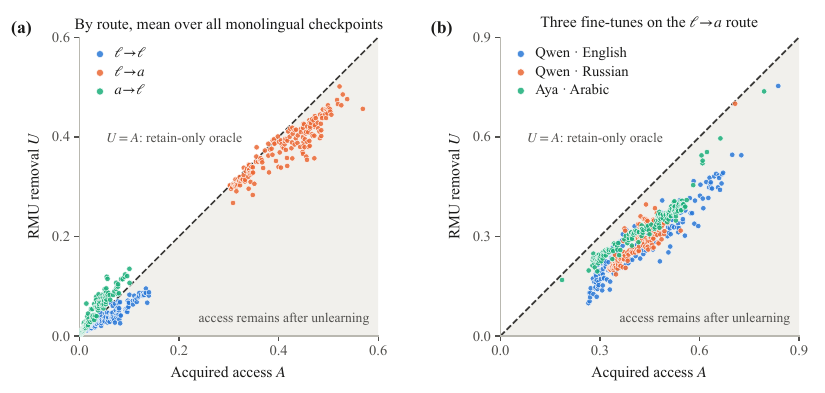}
\caption{\textbf{Uneven RMU transfer across language routes.} Points are evaluation languages; acquired access is $A=p_{\mathrm{FT}}-p_{\mathrm{O}}$ and removal is $U=p_{\mathrm{FT}}-p_{\mathrm{U}}$. Shading marks $U<A$. (a) Means over monolingual checkpoints. (b) Qwen English, Qwen Russian, and Aya Arabic fine-tunes on $\ell\!\to\!a$.}
\label{fig:findings-rmu-overlays}
\label{fig:findings-aggregate-transfer}
\label{fig:findings-selected-checkpoint-transfer}
\end{figure}

\paragraph{Does unlearning transfer across languages?}
The first question we might ask of the \benchmark is when we finetune in one language, to what extent does that knowledge become accessible in our other 173 language--script pairs? We see in \autoref{tab:transfer-acquisition-unlearning}A that uplift across languages is broad but concentrated on the query side. Gains are much smaller when the answer must also be expressed in another language. In \autoref{tab:acquisition-rankings}b and \autoref{tab:acquisition-rankings}c we rank languages by how much access they gain when only the query language is allowed to change and when both query and answer language change from the fine-tuned language.  In terms of acquisition languages that produce the largest mean gains across $\ell \rightarrow \ell$ routes, Spanish wins across all three model families \autoref{tab:acquisition-rankings}a.
\begin{table}[t]
\centering
\caption{\textbf{Acquisition and unlearning transfer across language routes.} (A) Gains over the base model after fine-tuning in $a$; sufficient access requires $p_{\mathrm{FT}}\geq0.10$ and gain $\geq0.05$. (B) Oracle-normalized removal after unlearning in $a$; final columns report languages reaching $\geq90\%$ removal and worst-language residual. Values above 100 indicate suppression below the retain-only oracle.}
\label{tab:transfer-acquisition-unlearning}
\label{tab:acquisition-transfer-base}
\label{tab:unlearning-transfer-panel-b}
\begingroup
\small
\setlength{\tabcolsep}{4pt}
\renewcommand{\arraystretch}{1.1}
\begin{tabular*}{\textwidth}{@{\extracolsep{\fill}}lrrrrr@{}}
\toprule
\multicolumn{6}{@{}l}{\textbf{(A) Acquisition: likelihood gain over base}} \\
\addlinespace[2pt]
Model & $a\!\to\!a$ & $\ell\!\to\!a$ & $\ell\!\to\!\ell$ & \makecell{$\ell\!\to\!\ell$ with\\sufficient access} & \\
\midrule
Aya   & $+0.85$ & $+0.49$ & $+0.14$ & 65.3\% & \\
Qwen  & $+0.84$ & $+0.47$ & $+0.10$ & 65.5\% & \\
Llama & $+0.76$ & $+0.45$ & $+0.03$ & 29.6\% & \\
Mean  & $+0.82$ & $+0.47$ & $+0.09$ & 53.5\% & \\
\midrule
\addlinespace[3pt]
\multicolumn{6}{@{}l}{\textbf{(B) Unlearning: oracle-normalized removal (\%)}} \\
\addlinespace[2pt]
Method & $a\!\to\!a$ & $\ell\!\to\!a$ & $\ell\!\to\!\ell$ & \makecell{Other languages\\$\geq 90\%$ removed} & \makecell{Worst language:\\access left} \\
\midrule
SimNPO   & 97.2  & 91.0 & 81.8  & 30.0 & 49.2 \\
GradDiff & 99.7  & 96.4 & 110.5 & 71.1 & 26.2 \\
RMU      & 100.2 & 91.1 & 63.5  & 18.0 & 75.4 \\
\bottomrule
\end{tabular*}
\endgroup
\end{table}

\paragraph{Do languages that show strong transfer during learning also show strong forgetting during unlearning?}
We might next wonder to what extent unlearning in one language transfers to others, and second, if there is a relationship between languages that learn from each other and languages that unlearn from each other. In \autoref{tab:transfer-acquisition-unlearning}B we see that unlearning leaves behind residual access in a large number of \textit{side-door languages}. Across the full tensor, less than 20\% of alternative access routes reach the same closure level as $a \rightarrow a$, suggesting a cross lingual asymmetry in learning vs unlearning. This asymmetry is well defined in \autoref{fig:findings-rmu-overlays}, where points below the diagonal show residual access remaining greater than the oracle matched acquisition language removal. In \autoref{tab:destination-language-bottlenecks} we rank the greatest language bottlenecks, languages with the worst unlearning transfer across all unlearning methods. The languages we see in this list are varied. The majority are low resource, but Russian and Chinese also consistently rank as tricky removal cases. In \autoref{tab:acquisition-removal-rank-agreement} we examine if languages that gain more are also the ones that lose more. We see that stronger acquisition transfer generally goes hand in hand with larger absolute drops. \autoref{fig:findings-rmu-overlays} illustrates this distinction for RMU. Points below the diagonal indicate that removal falls short of the acquired access above the retain-only reference. Panel (a) of \autoref{fig:findings-rmu-overlays} shows the large discrepancy in access when query vs answer is varied.
\paragraph{Does the way we ask matter?}
As a motivating experiment to see how various representations of the same fact affect access, we look at Romanized vs native scripts in a fixed language setting. Consider the setting where we unlearn with native Hindi questions vs. unlearning with Romanized Hindi questions holding native Hindi answers fixed. After unlearning with Romanized Hindi questions, we see weak removal when evaluating with native Hindi script (and vice versa). See \autoref{tab:hindi-script-intervention} for the full result. This suggests our evaluated unlearning methods suppress syntactic routes rather than semantic pathways, motivating our detailed examination of paraphrases. Accordingly, we report which of our paraphrase types leave the correct answer most accessible after unlearning. Modal verb paraphrases rank first in all three model families (\autoref{tab:expression-paraphrase-type-ranking}). Subordination and sentence-modality variants also rank highly, although their ordering varies across models. Qualitatively, we observe that the stronger paraphrase types often change how information is requested (\autoref{fig:form-examples}).
\begin{figure}[t]
\centering
\begin{minipage}{\textwidth}
{\small\textbf{(a)}\enspace\textit{Script:} unlearned with Romanized Hindi questions, native Hindi answers.}\par
\input{figs/findings/hindi_qualitative_example}
{\small\textbf{(b)}\enspace\textit{Phrasing:} two paraphrases of one English question, after unlearning.}\par
\input{figs/findings/paraphrase_qualitative_example}
\end{minipage}
\caption{\textbf{Residual access depends on expression.} (a) Gold-answer likelihood before and after unlearning; changing script leaves the fact accessible. (b) Post-unlearning gold-answer likelihood varies across paraphrases.}
\label{fig:form-examples}
\end{figure}

\section{Language budgeted unlearning with \method}
\label{sec:source-selection}
Our goal is to select $K$ source languages efficiently, without held out forget data at deployment, and generalize to unseen deletion requests.
\subsection{Method}
To achieve this goal, we measure how similarly the model represents calibration data across languages. The pairwise alignment between source and held out languages informs a candidate source set's coverage, or how well we anticipate supervision on a source set will generalize to the held out set. We  compare a fixed versus learned approach for scoring coverage. Fixed coverage uses cross-language similarities directly; learned coverage fits a scorer to historical unlearning outcomes. We draw inspiration from prior work that compares the similarity of semantically matched versus mismatched translated examples \citep{zeng-etal-2025-converging}. For each source to held out language pair we define a margin as an example's similarity to its direct translation minus its similarity to the most similar non matching example in the held out calibration set. We compute cosine similarities of final prompt token hidden states across decoder blocks, using two disjoint calibration sets $\mathcal{B}^{(p)}$, $p\in\{1,2\}$, drawn from the retain set to reduce sensitivity to individual samples.
For example $i$ in calibration set $p$ and language $\ell$, let $h_{\theta,b}(x_{i,\ell}^{(p)})$ denote decoder block $b$'s output at the final prompt token. Averaging over the $B_\theta$ decoder blocks gives
\(
a_{s,t}^{(p)}(i,j)=\frac{1}{B_\theta}\sum_{b=1}^{B_\theta}
\cos\!\left(h_{\theta,b}(x_{i,s}^{(p)}),h_{\theta,b}(x_{j,t}^{(p)})\right).
\)
Here $a_{s,t}^{(p)}(i,j)$ compares example $i$ in the source language with example $j$ in the target language. We define the per-example similarity margin as
$$
\delta_{s\rightarrow t}^{(p)}(i)
=
a_{s,t}^{(p)}(i,i)
-\max_{\substack{1\leq j\leq n_p\\j\neq i}}
a_{s,t}^{(p)}(i,j),
\qquad n_p=|\mathcal{B}^{(p)}|.
$$

Averaging margins within each calibration set and weighting the sets equally gives $M_{s\rightarrow t}=\frac12\sum_{p=1}^{2}\frac1{n_p}\sum_{i=1}^{n_p}\delta_{s\rightarrow t}^{(p)}(i)$ for $s\neq t$.

\paragraph{Fixed coverage}
For candidate source sets $\mathcal{S}_K=\{S\subseteq\Lall:|S|=K\}$, let $T(S)=\Lall\setminus S$. Fixed coverage averages each held-out language's strongest source margin:
$$ c_t(S)=\max_{s\in S}M_{s\rightarrow t},
\qquad
F_{\mathrm{FR}}(S)
=\frac{1}{|T(S)|}\sum_{t\in T(S)}c_t(S),
\qquad
\widehat S_{\mathrm{FR}}
=\operatorname*{arg\,max}_{S\in\mathcal{S}_K}F_{\mathrm{FR}}(S).
$$
\paragraph{Learned coverage} Inspired by LangRank \citep{lin-etal-2019-choosing}, which implements a learned ranker from transfer languages to downstream task languages, \method learns to rank source sets from historical unlearning outcomes. Our features capture weak or uneven target coverage and support from multiple sources with the mean, minimum, and standard deviation of $c_t(S)$ across held-out languages; the mean and minimum $M_{s\to t}$ over pairs of selected sources and held-out targets; and the target-averaged mean of each target's two strongest source margins. We standardize each feature using its historical training mean and standard deviation to obtain $\widetilde\phi(S)\in\mathbb{R}^6$. We fit a linear scorer with weighted pairwise logistic loss and $\ell_2$ regularization, comparing sets only within the same parent checkpoint, candidate inventory, and historical deletion request. The resulting score is $F_{\mathrm{PW}}(S)=w^\top\widetilde\phi(S)$, selecting $\widehat S_{\mathrm{PW}}=\operatorname{arg\,max}_{S\in\mathcal{S}_K}F_{\mathrm{PW}}(S)$. The learned ranker fits 180 subset outcomes from the nine development grids, each averaged over five orders.

\paragraph{Optional source language weighting}
\method uses uniform weighting in the forget loss for each source language. We test if after fixing the source set, additional weightings can also improve performance on held-out language forgetting. We again restrict ourselves to using only benign calibration data to estimate source to held out language gradient affinities and choose source weights that maximize the minimum weighted affinity to held out languages. The resulting weights scale only the forget loss. See \autoref{app:source-weighting} for full method details.

\subsection{Experiment design}

We evaluate Tiny Aya, Qwen3, and Llama (\autoref{app:checkpoints}) fine-tuned on three six-language sets. \emph{script-diverse} (EN, AR, JA, RU, ES, ZH), \emph{Latin-script} (EN, DE, ID, SW, TR, VI), and \emph{low-resource} (EN, AR, HA, HI, NE, SW). For each of the nine checkpoints, we run SimNPO with uniform source weighting on all 41 subsets of sizes one through three. Repeated rankings focus on all 20 triples ($K=3$). Endpoints follow \autoref{sec:protocol}; the primary metric is mean residual $R$ over native routes in the $6-K$ languages without forget supervision.

To generate empirical ranking we run every triple under five training data orderings. We rank the 20 triples separately within each order and summarize ranking agreement by the median pairwise Spearman correlation across orders. We also select the triple with lowest mean residual over four orders and evaluate it against uniform selection on the fifth, rotating across all five held-out orders. Across disjoint forget requests, we repeat the same grid and compute Spearman correlations between the order-averaged subset rankings. This tests whether useful rankings recur when the forget set changes.

We include as a baseline Individual top-3 which ranks six singleton sources by mean residual in the other five languages over three orders, then jointly unlearns on the three lowest residual sources. It requires 18 preliminary runs per model and target request data in all six languages; \method uses benign calibration at deployment. A deletion request specifies the facts to remove. We label two additional TOFU forget sets A and B; each contains 400 QA pairs and is disjoint from the other and the development forget set. These sets were excluded from fitting.

\subsection{Results}
\paragraph{Source selector performance.}
Our results support that selection choice matters. Across all checkpoints we observe that the best language subsets improve held out residual by 14.0-31.4\% over uniform. In the low resource inventory, the worst subsets leave 2.04$\times$, 1.94 $\times$ and 1.92 $\times$ the residual access of the best subsets for Aya, Llama, and Qwen. See \autoref{tab:development-ranking-consistency} for the full breakdown of all model family-language set pairs. This establishes substantial headroom in the best versus worst source sets, but do `good' empirical rankings generalize across disjoint forget requests? \autoref{tab:source-request-stability} shows the strong correlations both within requests where the orders vary and between disjoint sets. We see the weakest correlation in the Latin script inventory. Its smaller median pairwise gaps, 2.36-3.05 oracle normalized points versus 5.13-6.40 for script-diverse and 7.63-11.33 for low-resource, may make rankings harder to resolve. Overall, we observe that strong correlation does not guarantee a universal best subset. Qwen request B has cross-request \(\rho=0.668\), but only one shared top-five subset. This supports recurring ranking structure more strongly than a universally reusable best source set. \autoref{fig:findings-rank-heatmaps} illustrates this shared structure well.

\begin{table}[t]
\centering
\caption{\textbf{Source choice affects forgetting, with inventory-dependent ranking stability.} Residuals average five training orders; best and worst triples minimize and maximize this mean. Uniform averages all 20 triples. Gains are point or relative reductions from uniform. $\rho$ is the median pairwise Spearman correlation across order-specific rankings.}
\label{tab:development-ranking-consistency}
\begingroup
\small
\setlength{\tabcolsep}{4pt}
\renewcommand{\arraystretch}{1.1}
\begin{tabular*}{\textwidth}{@{\extracolsep{\fill}}llrrrrrrc@{}}
\toprule
 & & \multicolumn{3}{c}{Residual access $100R$ $\downarrow$} & \multicolumn{3}{c}{Gain over uniform $\uparrow$} & \makecell{Ranking\\consistency} \\
\cmidrule(lr){3-5}\cmidrule(lr){6-8}
Inventory & Model & Best & Uniform & Worst & Best (pts) & Worst (pts) & Best (\%) & $\rho$ \\
\midrule
Script-diverse & Qwen  & 33.59 & 41.94 & 52.93 & $+8.35$  & $-10.99$ & 19.9 & 0.813 \\
               & Aya   & 31.38 & 41.38 & 54.17 & $+10.00$ & $-12.79$ & 24.2 & 0.850 \\
               & Llama & 46.15 & 53.69 & 62.26 & $+7.53$  & $-8.57$  & 14.0 & 0.663 \\
\addlinespace
Low-resource   & Qwen  & 34.32 & 48.60 & 65.92 & $+14.28$ & $-17.32$ & 29.4 & 0.886 \\
               & Aya   & 28.89 & 39.85 & 58.97 & $+10.96$ & $-19.12$ & 27.5 & 0.875 \\
               & Llama & 33.40 & 48.71 & 64.96 & $+15.31$ & $-16.25$ & 31.4 & 0.805 \\
\addlinespace
Latin-script   & Qwen  & 28.02 & 32.99 & 41.48 & $+4.97$  & $-8.49$  & 15.1 & 0.408 \\
               & Aya   & 20.41 & 24.97 & 30.09 & $+4.56$  & $-5.12$  & 18.3 & 0.453 \\
               & Llama & 28.82 & 34.66 & 47.37 & $+5.84$  & $-12.71$ & 16.8 & 0.203 \\
\bottomrule
\end{tabular*}
\endgroup
\end{table}

\begin{table}[t]
\centering
\caption{\textbf{Source selection across deletion requests and datasets.} Gains are point reductions in $100R$ from uniform over all 20 triples; extra retain damage is in retained-probability points. (a) TOFU: mean $\pm$ SD over five orders; ``= Fixed'' marks identical subsets. (b) LUME request B: two realizations; all-six includes sources. (c) LORELEI: means over two orders; $\Delta$chrF++ is on retained generations. Historical best reuses the best subset from an earlier request.}
\label{tab:selector-generalization}
\label{tab:selector-tofu-requests}
\label{tab:selector-lume-request-b}
\label{tab:selector-lorelei}
\begingroup
\small
\setlength{\tabcolsep}{4pt}
\renewcommand{\arraystretch}{1.1}
\begin{tabular*}{\textwidth}{@{\extracolsep{\fill}}lrrrr@{}}
\toprule
\multicolumn{5}{@{}l}{\textbf{(a) TOFU, low-resource inventory, two held-out requests}} \\
\addlinespace[2pt]
 & & \multicolumn{3}{c}{Reduction in held-out $100R$ $\uparrow$} \\
\cmidrule(lr){3-5}
Model / request & Uniform $100R\downarrow$ & Fixed coverage & Historical best & \method \\
\midrule
Aya / A   & 39.23 & $+10.70$\sdv{1.62} & $+7.03$\sdv{2.43}  & = Fixed \\
Aya / B   & 36.57 & $+9.84$\sdv{0.89}  & $+9.36$\sdv{2.10}  & = Fixed \\
Qwen / A  & 49.94 & $+6.72$\sdv{3.13}  & $+5.65$\sdv{1.87}  & = Fixed \\
Qwen / B  & 47.06 & $+6.14$\sdv{4.29}  & $+4.38$\sdv{2.96}  & = Fixed \\
Llama / A & 48.66 & $+3.68$\sdv{3.84}  & $+11.25$\sdv{4.27} & $+3.80$\sdv{6.94} \\
Llama / B & 49.91 & $+6.18$\sdv{3.96}  & $+7.50$\sdv{4.03}  & $+7.48$\sdv{1.94} \\
\bottomrule
\end{tabular*}
\vspace{5pt}
\begin{tabular*}{\textwidth}{@{\extracolsep{\fill}}llcrrrr@{}}
\toprule
\multicolumn{7}{@{}l}{\textbf{(b) LUME, one held-out request}} \\
\addlinespace[2pt]
Selection policy & Sources & \makecell{Rank\\r1/r2} & \makecell{Held-out\\$100R\downarrow$} & \makecell{Held-out\\reduction $\uparrow$} & \makecell{All-six\\reduction $\uparrow$} & \makecell{Extra retain\\damage $\downarrow$} \\
\midrule
Uniform         & all 20 triples & --   & 40.84 & 0       & 0       & 0 \\
Fixed coverage  & JA/ES/ZH       & 7/10 & 39.26 & $+1.58$ & $+0.90$ & $-3.19$ \\
Historical best & AR/EN/JA       & 11/5 & 38.82 & $+2.02$ & $+1.41$ & $-1.75$ \\
\method         & EN/JA/RU       & 1/1  & 32.13 & $+8.71$ & $+4.76$ & $+1.30$ \\
\bottomrule
\end{tabular*}
\vspace{5pt}
\begin{tabular*}{\textwidth}{@{\extracolsep{\fill}}llrrrr@{}}
\toprule
\multicolumn{6}{@{}l}{\textbf{(c) LORELEI news documents with human translations}} \\
\addlinespace[2pt]
 & & \multicolumn{2}{c}{Held-out reduction $\uparrow$} & & \\
\cmidrule(lr){3-4}
Model & Policy & \makecell{Document\\beginning} & \makecell{Document\\interior} & \makecell{Extra retain\\damage $\downarrow$} & \makecell{Retained\\$\Delta$chrF++ $\uparrow$} \\
\midrule
Aya  & \method (= Fixed) & $+4.60$ & $+1.45$ & $-0.43$ & $-5.57$ \\
Qwen & \method (= Fixed) & $+2.31$ & $+0.79$ & $+0.75$ & $-1.75$ \\
\bottomrule
\end{tabular*}
\endgroup
\end{table}

We next consider if \method exploits the empirical headroom we observe on disjoint forget requests (\autoref{tab:selector-generalization}a). \method reduces mean residual access in held out languages without forget supervision by 7.8--27.3\% relative to uniform selection. It chooses the same subsets as fixed coverage for Aya and Qwen. These results support useful source selection across forget requests.

We evaluate Qwen3-4B on a new LUME document-removal request of 173 document groups, disjoint from the earlier request and calibration data (\autoref{tab:selector-generalization}b). Source choices are fixed before training. \method selects the subset with the lowest held out residual access in both training runs on this forget set. It reduces mean residual access by 21.3\% relative to uniform selection, compared with 3.9\% for fixed coverage and 4.9\% for reusing the best subset from the previous LUME forget request.

\paragraph{Behavioral and LORELEI  performance.}
\begin{table}[t]
\centering
\caption{\textbf{Generated-answer evaluation on the script-diverse inventory.} ES and behavior (\%), averaged over three orders and six native routes (retention: Aya six, Qwen/Llama five excluding Arabic). Random controls: EN/AR/ZH, EN/JA/ES, AR/JA/RU. Individual top-3 selects the strongest singleton sources. Metric definitions: \autoref{app:generation-judge}.}
\label{tab:selector-original-behavior}
\begingroup
\small
\setlength{\tabcolsep}{4pt}
\renewcommand{\arraystretch}{1.1}
\begin{tabular*}{\textwidth}{@{\extracolsep{\fill}}llrrrrrr@{}}
\toprule
Model & Policy & ES $\downarrow$ & \makecell{Full\\leakage $\downarrow$} & \makecell{Any\\disclosure $\downarrow$} & \makecell{Retained\\recovery $\uparrow$} & \makecell{Gibberish\\$\downarrow$} & \makecell{Code-\\switching $\downarrow$} \\
\midrule
Aya & Controls & 0.2572 & 29.35 & 52.28 & 69.48 & \textbf{1.52} & \textbf{6.70} \\
Aya & Individual top-3 & -- & 29.76 & 52.15 & 70.33 & 4.06 & 8.76 \\
Aya & \method & \textbf{0.2319} & \textbf{25.57} & \textbf{47.13} & \textbf{71.44} & 4.39 & 9.41 \\
\midrule
Qwen & Controls & 0.2433 & 27.32 & 50.93 & \textbf{75.24} & 2.41 & 7.92 \\
Qwen & Individual top-3 & -- & \textbf{25.22} & 49.63 & 72.93 & \textbf{1.72} & 8.22 \\
Qwen & \method & \textbf{0.2215} & \textbf{25.22} & \textbf{48.13} & 71.60 & 2.91 & \textbf{6.52} \\
\midrule
Llama & Controls & 0.2155 & 28.50 & 52.04 & 73.78 & 11.06 & 11.98 \\
Llama & Individual top-3 & -- & 32.37 & 56.70 & 74.13 & \textbf{7.11} & \textbf{10.89} \\
Llama & \method & \textbf{0.1940} & \textbf{25.72} & \textbf{50.33} & \textbf{79.60} & 9.31 & 12.87 \\\bottomrule
\end{tabular*}
\endgroup
\end{table}

Do \method's  gains translate to an actual reduction in leakage? \autoref{tab:selector-original-behavior} shows that they do with \method beating controls on extraction strength, full leakage, partial leakage. Gibberish scores remain low as well suggesting these gains are not coming at the cost of model collapse. Compared to Individual top-3 \method improves Llama's leakage from 32.37\% to 25.72\%, Aya's 29.76\% to 25.57\% and Qwen ties. Although  gibberish increases for all three families, the fraction of answers with no disclosure, no judged gibberish and only the requested language increases by 5.6 percentage points for Llama and 4.6 percentage points for Aya. We also evaluate source selection on the human translated news documents in LORELEI (\autoref{tab:selector-generalization}c). \method reduces mean residual access in held out languages by 6.1\% for Aya and 2.6\% for Qwen when prompts begin at the start of a document. These improvements occur under both training orders. Improvements are smaller for prompts drawn from document interiors, at 2.0\% and 0.9\%, respectively. These results extend the evidence for useful source selection to natural documents, with smaller gains and measurable retention tradeoffs.

\paragraph{Source weighting shows model dependent tradeoffs.}
Finally we compare calibrated and uniform weights on a TOFU deletion request with three paired training orders with the script diverse language set. We find in Qwen with EN, AR, ZH source set, calibrated weights reduce leakage across all six languages from 34.44\% to 25.65\%. Retain also improves from 80.56\% to 81.44\% and gibberish decreases from 0.63\% to 0.43\%. Full leakage in held out JA, RU and ES decreases from 39.00\% to 29.59\% over uniform weighting. For Llama, improvements in leakage come at a sharp utility cost. Full disclosure falls from 31.76\% to 26.33\%, but gibberish jumps from 5.69\% to 18.48\%.

\label{sec:source-selection-anchor}
\section{Related Work}
\label{sec:related-work}
\paragraph{LLM unlearning}
MU is formalized in \citet{unlearning:sp15} and addresses the issue of removing the influence of targeted training data without requiring full retraining. For LLMs, methods for unlearning take a variety of forms including gradient ascent based methods \citep{jang-etal-2023-knowledge}, preference optimization methods \citep{zhang2024negative, fan2025simplicity}, representation based methods \citep{li2024wmdp} and localization interventions \cite{jia2024wagle, wu-etal-2023-depn}.
\paragraph{Multilingual unlearning}
Early studies of multilingual unlearning include observations by \citet{lu-koehn-2025-learn} that misinformation can propagate across languages in LMs and findings by \citet{choi-etal-2024-cross} that unlearning in one language can be insufficient. Since then a few works have continued the study of multilingual unlearning \citep{lizzo2026evaluatingcrosslingualunlearningmultilingual,xiang2026multilingual,farashah-etal-2026-multilingual,hwang2025uncoveringpotentialrisksunlearning, hwang-etal-2026-knowledge}. Up until now though, these studies have only analyzed a small set of languages, typically high-medium resource. Some early methods have been developed to address these issues. LingTea changes the retention objective for cross-lingual unlearning \citep{choi-etal-2024-cross}. Others have tried localization for crosslingual generalization, selecting language-agnostic layers for multilingual erasure \citep{li2026layertargetedmultilingualknowledgeerasure}. This asks where an update might transfer; we ask which source languages cover held out languages under a fixed budget while preserving retained facts and generation.
\paragraph{Multilingual unlearning benchmarks}
TOFU is our \benchmark's primary English-only building block and it introduced controlled fictitious biography unlearning with forget, retain and world-fact sets \citep{maini2024tofutaskfictitiousunlearning}. \citet{farashah-etal-2026-multilingual} provide a multilingual extension of this dataset and \cite{savelli2025famefictionalactorsmultilingual} also created an unlearning task of fictitious actors with language specific targets in English, French, German, Italian, and Spanish. These works use a limited language set. Our \benchmark includes 174 languages. Finally, we also note some concurrent work \citep{shao2026beyond, hwang2026mu2benchmultilingualmachineunlearning}.

\section{Conclusion}
We introduced language budgeted multilingual unlearning, where forget supervision is available in only a small source-language set but erased facts must be removed in held out languages. \benchmark shows that successful forgetting in supervised languages does not rule out substantial residual access through other query and answer languages. \method improves held-out forgetting through coverage guided source selection. Source weighting results do not establish a generally superior calibration rule across models but show a promising direction for future work.

\section*{Acknowledgments}

This research is based upon work supported in part by the Office of the Director of National Intelligence (ODNI), Intelligence Advanced Research Projects Activity (IARPA), via 56000026C0019. The views and conclusions contained herein are those of the authors and should not be interpreted as necessarily representing the official policies, either expressed or implied, of ODNI, IARPA, or the U.S. Government. The U.S. Government is authorized to reproduce and distribute reprints for governmental purposes notwithstanding any copyright annotation therein.

\subsection*{AI use statement}

In this work, we used generative AI tools for generating synthetic data sets, assisting with translation, cleaning and reformatting datasets and implementing methods. We have not used generative AI tools for helping develop theoretical models or conceptual frameworks, formulating mathematical claims, providing critical ingredients for proving mathematical claims, assisting in the writing of proofs, proposing or refining hypotheses, designing or providing feedback on research  methodology or experiments, supporting qualitative and thematic data analysis, interpreting results. Additionally, we used generative AI tools for creating or modifying scientific figures or images, creating or editing software code, editing a research paper to improve readability, proposing a title or keywords for a research paper and writing assistance. We have reviewed all AI-assisted work. We reviewed LLM generated text to ensure it was consistent with our findings. We take responsibility for the final content of this work, including text, claims or artifacts produced with the aid of generative AI.

\subsection*{Ethics statement}
No ethics concerns.

\subsection*{Reproducibility statement}

\begingroup
\raggedright
The project site is \url{https://tskow99.github.io/crosslingual-unlearning-tensor/}. 

Code is available at \url{https://github.com/tskow99/crosslingual-unlearning-tensor}. Benchmark construction and translation procedures are described in \autoref{app:benchmark} and \autoref{app:translation}. \autoref{app:reproducibility} documents training, endpoint selection, calibration. \autoref{app:generation-judge} describes behavioral scoring.\par
\endgroup

\bibliography{iclr2027_conference}

@inproceedings{
xiang2026multilingual,
title={Multilingual Unlearning in {LLM}s: Transfer, Dynamics, and Reversibility},
author={Chaoyi Xiang and Olga Ohrimenko and Benjamin I. P. Rubinstein and Lea Frermann},
booktitle={Forty-third International Conference on Machine Learning},
year={2026},
url={https://openreview.net/forum?id=PQqbcFehs4}
}

@misc{zhang2024negative,
      title={Negative Preference Optimization: From Catastrophic Collapse to Effective Unlearning},
      author={Ruiqi Zhang and Licong Lin and Yu Bai and Song Mei},
      year={2024},
      eprint={2404.05868},
      archivePrefix={arXiv},
      primaryClass={cs.LG},
      url={https://arxiv.org/abs/2404.05868},
}

@inproceedings{
fan2025simplicity,
title={Simplicity Prevails: Rethinking Negative Preference Optimization for {LLM} Unlearning},
author={Chongyu Fan and Jiancheng Liu and Licong Lin and Jinghan Jia and Ruiqi Zhang and Song Mei and Sijia Liu},
booktitle={The Thirty-ninth Annual Conference on Neural Information Processing Systems},
year={2025},
url={https://openreview.net/forum?id=JbvSQm5h1l}
}

@misc{li2024wmdp,
      title={The WMDP Benchmark: Measuring and Reducing Malicious Use With Unlearning},
      author={Nathaniel Li and Alexander Pan and Anjali Gopal and Summer Yue and Daniel Berrios and Alice Gatti and Justin D. Li and Ann-Kathrin Dombrowski and Shashwat Goel and Long Phan and Gabriel Mukobi and Nathan Helm-Burger and Rassin Lababidi and Lennart Justen and Andrew B. Liu and Michael Chen and Isabelle Barrass and Oliver Zhang and Xiaoyuan Zhu and Rishub Tamirisa and Bhrugu Bharathi and Adam Khoja and Zhenqi Zhao and Ariel Herbert-Voss and Cort B. Breuer and Samuel Marks and Oam Patel and Andy Zou and Mantas Mazeika and Zifan Wang and Palash Oswal and Weiran Lin and Adam A. Hunt and Justin Tienken-Harder and Kevin Y. Shih and Kemper Talley and John Guan and Russell Kaplan and Ian Steneker and David Campbell and Brad Jokubaitis and Alex Levinson and Jean Wang and William Qian and Kallol Krishna Karmakar and Steven Basart and Stephen Fitz and Mindy Levine and Ponnurangam Kumaraguru and Uday Tupakula and Vijay Varadharajan and Ruoyu Wang and Yan Shoshitaishvili and Jimmy Ba and Kevin M. Esvelt and Alexandr Wang and Dan Hendrycks},
      year={2024},
      eprint={2403.03218},
      archivePrefix={arXiv},
      primaryClass={cs.LG},
      url={https://arxiv.org/abs/2403.03218},
}

@inproceedings{lee2023language, series={WWW ’23},
   title={Do Language Models Plagiarize?},
   url={http://dx.doi.org/10.1145/3543507.3583199},
   DOI={10.1145/3543507.3583199},
   booktitle={Proceedings of the ACM Web Conference 2023},
   publisher={ACM},
   author={Lee, Jooyoung and Le, Thai and Chen, Jinghui and Lee, Dongwon},
   year={2023},
   month=Apr, pages={3637–3647},
   collection={WWW ’23} }

@misc{chang2023speak,
      title={Speak, Memory: An Archaeology of Books Known to ChatGPT/GPT-4},
      author={Kent K. Chang and Mackenzie Cramer and Sandeep Soni and David Bamman},
      year={2023},
      eprint={2305.00118},
      archivePrefix={arXiv},
      primaryClass={cs.CL},
      url={https://arxiv.org/abs/2305.00118},
}

@misc{carlini2021extracting,
      title={Extracting Training Data from Large Language Models},
      author={Nicholas Carlini and Florian Tramer and Eric Wallace and Matthew Jagielski and Ariel Herbert-Voss and Katherine Lee and Adam Roberts and Tom Brown and Dawn Song and Ulfar Erlingsson and Alina Oprea and Colin Raffel},
      year={2021},
      eprint={2012.07805},
      archivePrefix={arXiv},
      primaryClass={cs.CR},
      url={https://arxiv.org/abs/2012.07805},
}

@misc{lukas2023analyzing,
      title={Analyzing Leakage of Personally Identifiable Information in Language Models},
      author={Nils Lukas and Ahmed Salem and Robert Sim and Shruti Tople and Lukas Wutschitz and Santiago Zanella-Béguelin},
      year={2023},
      eprint={2302.00539},
      archivePrefix={arXiv},
      primaryClass={cs.LG},
      url={https://arxiv.org/abs/2302.00539},
}

@misc{dorna2025openunlearningacceleratingllmunlearning,
      title={OpenUnlearning: Accelerating LLM Unlearning via Unified Benchmarking of Methods and Metrics},
      author={Vineeth Dorna and Anmol Mekala and Wenlong Zhao and Andrew McCallum and Zachary C. Lipton and J. Zico Kolter and Pratyush Maini},
      year={2025},
      eprint={2506.12618},
      archivePrefix={arXiv},
      primaryClass={cs.CL},
      url={https://arxiv.org/abs/2506.12618},
}

@misc{maini2024tofutaskfictitiousunlearning,
      title={TOFU: A Task of Fictitious Unlearning for LLMs},
      author={Pratyush Maini and Zhili Feng and Avi Schwarzschild and Zachary C. Lipton and J. Zico Kolter},
      year={2024},
      eprint={2401.06121},
      archivePrefix={arXiv},
      primaryClass={cs.LG},
      url={https://arxiv.org/abs/2401.06121},
}

@misc{lizzo2026evaluatingcrosslingualunlearningmultilingual,
      title={Evaluating Cross-Lingual Unlearning in Multilingual Language Models},
      author={Tyler Lizzo and Larry Heck},
      year={2026},
      eprint={2601.06675},
      archivePrefix={arXiv},
      primaryClass={cs.CL},
      url={https://arxiv.org/abs/2601.06675},
}

@misc{li2026layertargetedmultilingualknowledgeerasure,
      title={Layer-Targeted Multilingual Knowledge Erasure in Large Language Models},
      author={Taoran Li and Varun Chandrasekaran and Zhiyuan Yu},
      year={2026},
      eprint={2602.22562},
      archivePrefix={arXiv},
      primaryClass={cs.CR},
      url={https://arxiv.org/abs/2602.22562},
}

@inproceedings{farashah-etal-2026-multilingual,
    title = "Multilingual Amnesia: On the Transferability of Unlearning in Multilingual {LLM}s",
    author = "Farashah, Alireza Dehghanpour  and
      Khandelwal, Aditi  and
      Fauchard, Marylou  and
      Shi, Zhuan  and
      Rostamzadeh, Negar  and
      Farnadi, Golnoosh",
    editor = "Demberg, Vera  and
      Inui, Kentaro  and
      Marquez, Llu{\'i}s",
    booktitle = "Proceedings of the 19th Conference of the {E}uropean Chapter of the {A}ssociation for {C}omputational {L}inguistics (Volume 1: Long Papers)",
    month = mar,
    year = "2026",
    address = "Rabat, Morocco",
    publisher = "Association for Computational Linguistics",
    url = "https://aclanthology.org/2026.eacl-long.260/",
    doi = "10.18653/v1/2026.eacl-long.260",
    pages = "5570--5589",
    ISBN = "979-8-89176-380-7"
}

@misc{hwang2025uncoveringpotentialrisksunlearning,
      title={Uncovering the Potential Risks in Unlearning: Danger of English-only Unlearning in Multilingual LLMs},
      author={Kyomin Hwang and Hyeonjin Kim and Seungyeon Kim and Sunghyun Wee and Nojun Kwak},
      year={2025},
      eprint={2510.23949},
      archivePrefix={arXiv},
      primaryClass={cs.CL},
      url={https://arxiv.org/abs/2510.23949},
}

@misc{savelli2025famefictionalactorsmultilingual,
      title={FAME: Fictional Actors for Multilingual Erasure},
      author={Claudio Savelli and Moreno La Quatra and Alkis Koudounas and Flavio Giobergia},
      year={2025},
      eprint={2512.15235},
      archivePrefix={arXiv},
      primaryClass={cs.CL},
      url={https://arxiv.org/abs/2512.15235},
}

@inproceedings{
jia2024wagle,
title={{WAGLE}: Strategic Weight Attribution for Effective and Modular Unlearning in Large Language Models},
author={Jinghan Jia and Jiancheng Liu and Yihua Zhang and Parikshit Ram and Nathalie Baracaldo and Sijia Liu},
booktitle={The Thirty-eighth Annual Conference on Neural Information Processing Systems},
year={2024},
url={https://openreview.net/forum?id=VzOgnDJMgh}
}

@inproceedings{lee-etal-2025-localization,
    title = "Does Localization Inform Unlearning? A Rigorous Examination of Local Parameter Attribution for Knowledge Unlearning in Language Models",
    author = "Lee, Hwiyeong  and
      Hwang, Uiji  and
      Lim, Hyelim  and
      Kim, Taeuk",
    editor = "Christodoulopoulos, Christos  and
      Chakraborty, Tanmoy  and
      Rose, Carolyn  and
      Peng, Violet",
    booktitle = "Proceedings of the 2025 Conference on Empirical Methods in Natural Language Processing",
    month = nov,
    year = "2025",
    address = "Suzhou, China",
    publisher = "Association for Computational Linguistics",
    url = "https://aclanthology.org/2025.emnlp-main.1109/",
    doi = "10.18653/v1/2025.emnlp-main.1109",
    pages = "21857--21869",
    ISBN = "979-8-89176-332-6"
}

@inproceedings{kocmi-etal-2024-findings,
    title = "Findings of the {WMT}24 General Machine Translation Shared Task: The {LLM} Era Is Here but {MT} Is Not Solved Yet",
    author = "Kocmi, Tom  and
      Avramidis, Eleftherios  and
      Bawden, Rachel  and
      Bojar, Ond{\v{r}}ej  and
      Dvorkovich, Anton  and
      Federmann, Christian  and
      Fishel, Mark  and
      Freitag, Markus  and
      Gowda, Thamme  and
      Grundkiewicz, Roman  and
      Haddow, Barry  and
      Karpinska, Marzena  and
      Koehn, Philipp  and
      Marie, Benjamin  and
      Monz, Christof  and
      Murray, Kenton  and
      Nagata, Masaaki  and
      Popel, Martin  and
      Popovi{\'c}, Maja  and
      Shmatova, Mariya  and
      Steingr{\'i}msson, Steinth{\'o}r  and
      Zouhar, Vil{\'e}m",
    editor = "Haddow, Barry  and
      Kocmi, Tom  and
      Koehn, Philipp  and
      Monz, Christof",
    booktitle = "Proceedings of the Ninth Conference on Machine Translation",
    month = nov,
    year = "2024",
    address = "Miami, Florida, USA",
    publisher = "Association for Computational Linguistics",
    url = "https://aclanthology.org/2024.wmt-1.1/",
    doi = "10.18653/v1/2024.wmt-1.1",
    pages = "1--46"
}

@article{guerreiro-etal-2023-hallucinations,
    title = "Hallucinations in Large Multilingual Translation Models",
    author = "Guerreiro, Nuno M.  and
      Alves, Duarte M.  and
      Waldendorf, Jonas  and
      Haddow, Barry  and
      Birch, Alexandra  and
      Colombo, Pierre  and
      Martins, Andr{\'e} F. T.",
    journal = "Transactions of the Association for Computational Linguistics",
    volume = "11",
    year = "2023",
    address = "Cambridge, MA",
    publisher = "MIT Press",
    url = "https://aclanthology.org/2023.tacl-1.85/",
    doi = "10.1162/tacl_a_00615",
    pages = "1500--1517"
}

@inproceedings{rei-etal-2022-cometkiwi,
    title = "{C}omet{K}iwi: {IST}-Unbabel 2022 Submission for the Quality Estimation Shared Task",
    author = "Rei, Ricardo  and
      Treviso, Marcos  and
      Guerreiro, Nuno M.  and
      Zerva, Chrysoula  and
      Farinha, Ana C  and
      Maroti, Christine  and
      C. de Souza, Jos{\'e} G.  and
      Glushkova, Taisiya  and
      Alves, Duarte  and
      Coheur, Luisa  and
      Lavie, Alon  and
      Martins, Andr{\'e} F. T.",
    editor = {Koehn, Philipp  and
      Barrault, Lo{\"i}c  and
      Bojar, Ond{\v{r}}ej  and
      Bougares, Fethi  and
      Chatterjee, Rajen  and
      Costa-juss{\`a}, Marta R.  and
      Federmann, Christian  and
      Fishel, Mark  and
      Fraser, Alexander  and
      Freitag, Markus  and
      Graham, Yvette  and
      Grundkiewicz, Roman  and
      Guzman, Paco  and
      Haddow, Barry  and
      Huck, Matthias  and
      Jimeno Yepes, Antonio  and
      Kocmi, Tom  and
      Martins, Andr{\'e}  and
      Morishita, Makoto  and
      Monz, Christof  and
      Nagata, Masaaki  and
      Nakazawa, Toshiaki  and
      Negri, Matteo  and
      N{\'e}v{\'e}ol, Aur{\'e}lie  and
      Neves, Mariana  and
      Popel, Martin  and
      Turchi, Marco  and
      Zampieri, Marcos},
    booktitle = "Proceedings of the Seventh Conference on Machine Translation (WMT)",
    month = dec,
    year = "2022",
    address = "Abu Dhabi, United Arab Emirates (Hybrid)",
    publisher = "Association for Computational Linguistics",
    url = "https://aclanthology.org/2022.wmt-1.60/",
    doi = "10.18653/v1/2022.wmt-1.60",
    pages = "634--645"
}

@inproceedings{ramakrishna-etal-2025-lume,
    title = "{LUME}: {LLM} Unlearning with Multitask Evaluations",
    author = "Ramakrishna, Anil  and
      Wan, Yixin  and
      Jin, Xiaomeng  and
      Chang, Kai-Wei  and
      Bu, Zhiqi  and
      Vinzamuri, Bhanukiran  and
      Cevher, Volkan  and
      Hong, Mingyi  and
      Gupta, Rahul",
    editor = "Christodoulopoulos, Christos  and
      Chakraborty, Tanmoy  and
      Rose, Carolyn  and
      Peng, Violet",
    booktitle = "Findings of the Association for Computational Linguistics: EMNLP 2025",
    month = nov,
    year = "2025",
    address = "Suzhou, China",
    publisher = "Association for Computational Linguistics",
    url = "https://aclanthology.org/2025.findings-emnlp.347/",
    doi = "10.18653/v1/2025.findings-emnlp.347",
    pages = "6524--6535",
    ISBN = "979-8-89176-335-7"
}

@misc{nllbteam2022languageleftbehindscaling,
      title={No Language Left Behind: Scaling Human-Centered Machine Translation},
      author={NLLB Team and Marta R. Costa-jussà and James Cross and Onur Çelebi and Maha Elbayad and Kenneth Heafield and Kevin Heffernan and Elahe Kalbassi and Janice Lam and Daniel Licht and Jean Maillard and Anna Sun and Skyler Wang and Guillaume Wenzek and Al Youngblood and Bapi Akula and Loic Barrault and Gabriel Mejia Gonzalez and Prangthip Hansanti and John Hoffman and Semarley Jarrett and Kaushik Ram Sadagopan and Dirk Rowe and Shannon Spruit and Chau Tran and Pierre Andrews and Necip Fazil Ayan and Shruti Bhosale and Sergey Edunov and Angela Fan and Cynthia Gao and Vedanuj Goswami and Francisco Guzmán and Philipp Koehn and Alexandre Mourachko and Christophe Ropers and Safiyyah Saleem and Holger Schwenk and Jeff Wang},
      year={2022},
      eprint={2207.04672},
      archivePrefix={arXiv},
      primaryClass={cs.CL},
      url={https://arxiv.org/abs/2207.04672},
}

@inproceedings{lin-etal-2019-choosing,
    title = "Choosing Transfer Languages for Cross-Lingual Learning",
    author = "Lin, Yu-Hsiang  and
      Chen, Chian-Yu  and
      Lee, Jean  and
      Li, Zirui  and
      Zhang, Yuyan  and
      Xia, Mengzhou  and
      Rijhwani, Shruti  and
      He, Junxian  and
      Zhang, Zhisong  and
      Ma, Xuezhe  and
      Anastasopoulos, Antonios  and
      Littell, Patrick  and
      Neubig, Graham",
    editor = "Korhonen, Anna  and
      Traum, David  and
      M{\`a}rquez, Llu{\'i}s",
    booktitle = "Proceedings of the 57th Annual Meeting of the Association for Computational Linguistics",
    month = jul,
    year = "2019",
    address = "Florence, Italy",
    publisher = "Association for Computational Linguistics",
    url = "https://aclanthology.org/P19-1301/",
    doi = "10.18653/v1/P19-1301",
    pages = "3125--3135"
}

@misc{salamanca2026tinyayabridgingscale,
      title={Tiny Aya: Bridging Scale and Multilingual Depth},
      author={Alejandro R. Salamanca and Diana Abagyan and Daniel D'souza and Ammar Khairi and David Mora and Saurabh Dash and Viraat Aryabumi and Sara Rajaee and Mehrnaz Mofakhami and Ananya Sahu and Thomas Euyang and Brittawnya Prince and Madeline Smith and Hangyu Lin and Acyr Locatelli and Sara Hooker and Tom Kocmi and Aidan Gomez and Ivan Zhang and Phil Blunsom and Nick Frosst and Joelle Pineau and Beyza Ermis and Ahmet Üstün and Julia Kreutzer and Marzieh Fadaee},
      year={2026},
      eprint={2603.11510},
      archivePrefix={arXiv},
      primaryClass={cs.CL},
      url={https://arxiv.org/abs/2603.11510},
}

@misc{yang2025qwen3technicalreport,
      title={Qwen3 Technical Report},
      author={An Yang and Anfeng Li and Baosong Yang and Beichen Zhang and Binyuan Hui and Bo Zheng and Bowen Yu and Chang Gao and Chengen Huang and Chenxu Lv and Chujie Zheng and Dayiheng Liu and Fan Zhou and Fei Huang and Feng Hu and Hao Ge and Haoran Wei and Huan Lin and Jialong Tang and Jian Yang and Jianhong Tu and Jianwei Zhang and Jianxin Yang and Jiaxi Yang and Jing Zhou and Jingren Zhou and Junyang Lin and Kai Dang and Keqin Bao and Kexin Yang and Le Yu and Lianghao Deng and Mei Li and Mingfeng Xue and Mingze Li and Pei Zhang and Peng Wang and Qin Zhu and Rui Men and Ruize Gao and Shixuan Liu and Shuang Luo and Tianhao Li and Tianyi Tang and Wenbiao Yin and Xingzhang Ren and Xinyu Wang and Xinyu Zhang and Xuancheng Ren and Yang Fan and Yang Su and Yichang Zhang and Yinger Zhang and Yu Wan and Yuqiong Liu and Zekun Wang and Zeyu Cui and Zhenru Zhang and Zhipeng Zhou and Zihan Qiu},
      year={2025},
      eprint={2505.09388},
      archivePrefix={arXiv},
      primaryClass={cs.CL},
      url={https://arxiv.org/abs/2505.09388},
}

@inproceedings{zhao-etal-2024-tracing,
    title = "Tracing the Roots of Facts in Multilingual Language Models: Independent, Shared, and Transferred Knowledge",
    author = "Zhao, Xin  and
      Yoshinaga, Naoki  and
      Oba, Daisuke",
    editor = "Graham, Yvette  and
      Purver, Matthew",
    booktitle = "Proceedings of the 18th Conference of the European Chapter of the Association for Computational Linguistics (Volume 1: Long Papers)",
    month = mar,
    year = "2024",
    address = "St. Julian{'}s, Malta",
    publisher = "Association for Computational Linguistics",
    url = "https://aclanthology.org/2024.eacl-long.127/",
    doi = "10.18653/v1/2024.eacl-long.127",
    pages = "2088--2102"
}

@inproceedings{qi-etal-2023-cross,
    title = "Cross-Lingual Consistency of Factual Knowledge in Multilingual Language Models",
    author = "Qi, Jirui  and
      Fern{\'a}ndez, Raquel  and
      Bisazza, Arianna",
    editor = "Bouamor, Houda  and
      Pino, Juan  and
      Bali, Kalika",
    booktitle = "Proceedings of the 2023 Conference on Empirical Methods in Natural Language Processing",
    month = dec,
    year = "2023",
    address = "Singapore",
    publisher = "Association for Computational Linguistics",
    url = "https://aclanthology.org/2023.emnlp-main.658/",
    doi = "10.18653/v1/2023.emnlp-main.658",
    pages = "10650--10666"
}

@inproceedings{unlearning:sp15,
    author = "Cao, Yinzhi and Yang, Junfeng",
    title = "Towards Making Systems Forget with Machine Unlearning",
    booktitle = "Proceedings of the 36th IEEE Symposium on Security and Privacy (IEEE S\&P)",
    shortbooktitle = "IEEE S\\&P",
    doi = "10.1109/SP.2015.35",
    pdfurl = "https://www.ieee-security.org/TC/SP2015/papers-archived/6949a463.pdf",
    pages = "463--480",
    year = "2015",
    month = "May"
}

@inproceedings{kovatchev-etal-2018-etpc,
    title = "{ETPC} - A Paraphrase Identification Corpus Annotated with Extended Paraphrase Typology and Negation",
    author = "Kovatchev, Venelin  and
      Mart{\'i}, M. Ant{\`o}nia  and
      Salam{\'o}, Maria",
    editor = "Calzolari, Nicoletta  and
      Choukri, Khalid  and
      Cieri, Christopher  and
      Declerck, Thierry  and
      Goggi, Sara  and
      Hasida, Koiti  and
      Isahara, Hitoshi  and
      Maegaard, Bente  and
      Mariani, Joseph  and
      Mazo, H{\'e}l{\`e}ne  and
      Moreno, Asuncion  and
      Odijk, Jan  and
      Piperidis, Stelios  and
      Tokunaga, Takenobu",
    booktitle = "Proceedings of the Eleventh International Conference on Language Resources and Evaluation ({LREC} 2018)",
    month = may,
    year = "2018",
    address = "Miyazaki, Japan",
    publisher = "European Language Resources Association (ELRA)",
    url = "https://aclanthology.org/L18-1221/"
}

@inproceedings{strassel-tracey-2016-lorelei,
    title = "{LORELEI} Language Packs: Data, Tools, and Resources for Technology Development in Low Resource Languages",
    author = "Strassel, Stephanie  and
      Tracey, Jennifer",
    editor = "Calzolari, Nicoletta  and
      Choukri, Khalid  and
      Declerck, Thierry  and
      Goggi, Sara  and
      Grobelnik, Marko  and
      Maegaard, Bente  and
      Mariani, Joseph  and
      Mazo, Helene  and
      Moreno, Asuncion  and
      Odijk, Jan  and
      Piperidis, Stelios",
    booktitle = "Proceedings of the Tenth International Conference on Language Resources and Evaluation ({LREC}'16)",
    month = may,
    year = "2016",
    address = "Portoro{\v{z}}, Slovenia",
    publisher = "European Language Resources Association (ELRA)",
    url = "https://aclanthology.org/L16-1521/",
    pages = "3273--3280"
}

@misc{grattafiori2024llama3herdmodels,
      title={The Llama 3 Herd of Models},
      author={Aaron Grattafiori and Abhimanyu Dubey and Abhinav Jauhri and others},
      year={2024},
      eprint={2407.21783},
      archivePrefix={arXiv},
      primaryClass={cs.AI},
      url={https://arxiv.org/abs/2407.21783},
}

@inproceedings{zeng-etal-2025-converging,
    title = "Converging to a Lingua Franca: Evolution of Linguistic Regions and Semantics Alignment in Multilingual Large Language Models",
    author = "Zeng, Hongchuan  and
      Han, Senyu  and
      Chen, Lu  and
      Yu, Kai",
    editor = "Rambow, Owen  and
      Wanner, Leo  and
      Apidianaki, Marianna  and
      Al-Khalifa, Hend  and
      Eugenio, Barbara Di  and
      Schockaert, Steven",
    booktitle = "Proceedings of the 31st International Conference on Computational Linguistics",
    month = jan,
    year = "2025",
    address = "Abu Dhabi, UAE",
    publisher = "Association for Computational Linguistics",
    url = "https://aclanthology.org/2025.coling-main.707/",
    pages = "10602--10617"
}

@inproceedings{wu-etal-2023-depn,
    title = "{DEPN}: Detecting and Editing Privacy Neurons in Pretrained Language Models",
    author = "Wu, Xinwei  and
      Li, Junzhuo  and
      Xu, Minghui  and
      Dong, Weilong  and
      Wu, Shuangzhi  and
      Bian, Chao  and
      Xiong, Deyi",
    editor = "Bouamor, Houda  and
      Pino, Juan  and
      Bali, Kalika",
    booktitle = "Proceedings of the 2023 Conference on Empirical Methods in Natural Language Processing",
    month = dec,
    year = "2023",
    address = "Singapore",
    publisher = "Association for Computational Linguistics",
    url = "https://aclanthology.org/2023.emnlp-main.174/",
    doi = "10.18653/v1/2023.emnlp-main.174",
    pages = "2875--2886"
}

@inproceedings{bandarkar-etal-2024-belebele,
    title = "The Belebele Benchmark: a Parallel Reading Comprehension Dataset in 122 Language Variants",
    author = "Bandarkar, Lucas  and
      Liang, Davis  and
      Muller, Benjamin  and
      Artetxe, Mikel  and
      Shukla, Satya Narayan  and
      Husa, Donald  and
      Goyal, Naman  and
      Krishnan, Abhinandan  and
      Zettlemoyer, Luke  and
      Khabsa, Madian",
    editor = "Ku, Lun-Wei  and
      Martins, Andre  and
      Srikumar, Vivek",
    booktitle = "Proceedings of the 62nd Annual Meeting of the Association for Computational Linguistics (Volume 1: Long Papers)",
    month = aug,
    year = "2024",
    address = "Bangkok, Thailand",
    publisher = "Association for Computational Linguistics",
    url = "https://aclanthology.org/2024.acl-long.44/",
    doi = "10.18653/v1/2024.acl-long.44",
    pages = "749--775"
}

@inproceedings{singh-etal-2025-global,
    title = "Global {MMLU}: Understanding and Addressing Cultural and Linguistic Biases in Multilingual Evaluation",
    author = "Singh, Shivalika  and
      Romanou, Angelika  and
      Fourrier, Cl{\'e}mentine  and
      Adelani, David Ifeoluwa  and
      Ngui, Jian Gang  and
      Vila-Suero, Daniel  and
      Limkonchotiwat, Peerat  and
      Marchisio, Kelly  and
      Leong, Wei Qi  and
      Susanto, Yosephine  and
      Ng, Raymond  and
      Longpre, Shayne  and
      Ruder, Sebastian  and
      Ko, Wei-Yin  and
      Bosselut, Antoine  and
      Oh, Alice  and
      Martins, Andre  and
      Choshen, Leshem  and
      Ippolito, Daphne  and
      Ferrante, Enzo  and
      Fadaee, Marzieh  and
      Ermis, Beyza  and
      Hooker, Sara",
    editor = "Che, Wanxiang  and
      Nabende, Joyce  and
      Shutova, Ekaterina  and
      Pilehvar, Mohammad Taher",
    booktitle = "Proceedings of the 63rd Annual Meeting of the Association for Computational Linguistics (Volume 1: Long Papers)",
    month = jul,
    year = "2025",
    address = "Vienna, Austria",
    publisher = "Association for Computational Linguistics",
    url = "https://aclanthology.org/2025.acl-long.919/",
    doi = "10.18653/v1/2025.acl-long.919",
    pages = "18761--18799",
    ISBN = "979-8-89176-251-0"
}

@inproceedings{jang-etal-2023-knowledge,
    title = "Knowledge Unlearning for Mitigating Privacy Risks in Language Models",
    author = "Jang, Joel  and
      Yoon, Dongkeun  and
      Yang, Sohee  and
      Cha, Sungmin  and
      Lee, Moontae  and
      Logeswaran, Lajanugen  and
      Seo, Minjoon",
    editor = "Rogers, Anna  and
      Boyd-Graber, Jordan  and
      Okazaki, Naoaki",
    booktitle = "Proceedings of the 61st Annual Meeting of the Association for Computational Linguistics (Volume 1: Long Papers)",
    month = jul,
    year = "2023",
    address = "Toronto, Canada",
    publisher = "Association for Computational Linguistics",
    url = "https://aclanthology.org/2023.acl-long.805/",
    doi = "10.18653/v1/2023.acl-long.805",
    pages = "14389--14408"
}

@inproceedings{wang-etal-2024-afrimte,
    title = "{A}fri{MTE} and {A}fri{COMET}: Enhancing {COMET} to Embrace Under-resourced {A}frican Languages",
    author = "Wang, Jiayi  and
      Adelani, David Ifeoluwa  and
      Agrawal, Sweta  and others",
    editor = "Duh, Kevin  and
      Gomez, Helena  and
      Bethard, Steven",
    booktitle = "Proceedings of the 2024 Conference of the North American Chapter of the Association for Computational Linguistics: Human Language Technologies (Volume 1: Long Papers)",
    month = jun,
    year = "2024",
    address = "Mexico City, Mexico",
    publisher = "Association for Computational Linguistics",
    url = "https://aclanthology.org/2024.naacl-long.334/",
    doi = "10.18653/v1/2024.naacl-long.334",
    pages = "5997--6023"
}

@inproceedings{hwang-etal-2026-knowledge,
    title = "Knowledge Beyond Language: Bridging the Gap in Multilingual Machine Unlearning Evaluation",
    author = "Hwang, Kyomin  and
      Kim, Hyeonjin  and
      Cho, Sangyeon  and
      Kwak, Nojun",
    editor = "Liakata, Maria  and
      Moreira, Viviane P.  and
      Zhang, Jiajun  and
      Jurgens, David",
    booktitle = "Proceedings of the 64th Annual Meeting of the {A}ssociation for {C}omputational {L}inguistics (Volume 1: Long Papers)",
    month = jul,
    year = "2026",
    address = "San Diego, California, United States",
    publisher = "Association for Computational Linguistics",
    url = "https://aclanthology.org/2026.acl-long.1105/",
    doi = "10.18653/v1/2026.acl-long.1105",
    pages = "24085--24119",
    ISBN = "979-8-89176-390-6"
}

@inproceedings{popovic2017chrf++,
  title={chrF++: words helping character n-grams},
  author={Popovi{\'c}, Maja},
  booktitle={Proceedings of the second conference on machine translation},
  pages={612--618},
  year={2017}
}

@inproceedings{choi-etal-2024-cross,
    title = "Cross-Lingual Unlearning of Selective Knowledge in Multilingual Language Models",
    author = "Choi, Minseok  and
      Min, Kyunghyun  and
      Choo, Jaegul",
    editor = "Al-Onaizan, Yaser  and
      Bansal, Mohit  and
      Chen, Yun-Nung",
    booktitle = "Findings of the Association for Computational Linguistics: EMNLP 2024",
    month = nov,
    year = "2024",
    address = "Miami, Florida, USA",
    publisher = "Association for Computational Linguistics",
    url = "https://aclanthology.org/2024.findings-emnlp.630/",
    doi = "10.18653/v1/2024.findings-emnlp.630",
    pages = "10732--10747"
}

@inproceedings{lu-koehn-2025-learn,
    title = "Learn and Unlearn: Addressing Misinformation in Multilingual {LLM}s",
    author = "Lu, TaiMing  and
      Koehn, Philipp",
    editor = "Christodoulopoulos, Christos  and
      Chakraborty, Tanmoy  and
      Rose, Carolyn  and
      Peng, Violet",
    booktitle = "Proceedings of the 2025 Conference on Empirical Methods in Natural Language Processing",
    month = nov,
    year = "2025",
    address = "Suzhou, China",
    publisher = "Association for Computational Linguistics",
    url = "https://aclanthology.org/2025.emnlp-main.516/",
    doi = "10.18653/v1/2025.emnlp-main.516",
    pages = "10180--10195",
    ISBN = "979-8-89176-332-6"
}

@article{shao2026beyond,
  title={Beyond Cross-Lingual Transfer: Benchmarking Propagation Boundaries in Multilingual LLM Unlearning},
  author={Shao, Pengyang and Lu, Chuanpeng and Qin, Wei and Jin, Yanzheng and Liu, Xiaohao and Ai, Xi and Kawaguchi, Kenji and Hong, Richang},
  journal={arXiv preprint arXiv:2609.05976},
  year={2026}
}

@misc{hwang2026mu2benchmultilingualmachineunlearning,
      title={$\mu^2$-Bench: A Multilingual Machine Unlearning Benchmark},
      author={Kyomin Hwang and Hyeonjin Kim and Hyunho Lee and Yearim Kim and Yeji Song and Nojun Kwak},
      year={2026},
      eprint={2609.20945},
      archivePrefix={arXiv},
      primaryClass={cs.CL},
      url={https://arxiv.org/abs/2609.20945},
}
\bibliographystyle{paper_references}

\clearpage
\appendix
\raggedbottom
\section*{Appendix contents}
\label{app:contents}
\pdfbookmark[1]{Appendix contents}{app:contents-bookmark}

The appendix follows the order of the main text and is grouped into four parts. The
\emph{Supports} column gives the main-text section that each appendix section backs up; all entries
are links.

\vspace{6pt}
\begin{tabularx}{\linewidth}{@{}p{1.1em}Y>{\raggedleft\arraybackslash}p{5.2em}>{\raggedleft\arraybackslash}p{2.4em}@{}}
\toprule
 & & \textbf{Supports} & \textbf{Page}\\
\midrule
\apppart{I}{The benchmark}
\appsec{app:benchmark}{\S\ref{sec:protocol}, \ref{sec:benchmark}}{\appsub{app:source-data}\appsep\appsub{app:languages}\appsep\appsub{app:paraphrase}\appsep
  \appsub{app:checkpoints}\appsep\appsub{app:model-capability}\appsep
  \appsub{app:generation-judge}}
\appsec{app:translation}{\S\ref{sec:benchmark}}{\appsub{app:translation-engine}\appsep\appsub{app:translation-prompt}\appsep
  \appsub{app:translation-quality}\appsep\appsub{app:translation-screen}}
\apppart{II}{Supplementary benchmark findings}
\appsec{app:findings}{\S\ref{sec:findings}}{\appsub{app:findings-transfer-summary}\appsep\appsub{app:findings-unlearning-transfer}\appsep
  \appsub{app:findings-transfer-agreement}\appsep\appsub{app:findings-expression}\appsep
  \appsub{app:findings-form-policy}}
\apppart{III}{Language budgeted unlearning with \method}
\appsec{app:source-selection}{\S\ref{sec:source-selection}}{\appsub{app:all-language-supervision}\appsep\appsub{app:headroom-all-budgets}\appsep
  \appsub{app:source-request-stability}\appsep\appsub{app:selector-behavior}\appsep
  \appsub{app:source-weighting}\appsep
  \appsub{app:source-rankings}}
\apppart{IV}{Reproducibility}
\appsec{app:reproducibility}
  {\S\ref{sec:protocol}, \ref{sec:source-selection}}{}
\bottomrule
\end{tabularx}

\clearpage
\section{Benchmark construction}
\label{app:benchmark}
\label{app:protocol}
This appendix details how \benchmark is constructed (\autoref{sec:benchmark}).

\subsection{Source data}
\label{app:source-data}

Our English source data comes from TOFU \citep{maini2024tofutaskfictitiousunlearning} and LUME \citep{ramakrishna-etal-2025-lume}.

\subsection{Languages}
\label{app:languages}
Our inventory covers 174 NLLB language--script pairs spanning 18 scripts and three resource tiers (27 high, 46 medium, 101 low). \autoref{tab:language-list} provides the complete language list. Languages in green have translated paraphrase variants available.

\begin{table}[htbp]
\centering
\begin{threeparttable}
\caption{Language coverage and paraphrases. The benchmark
contains 174 language--script pairs spanning 18 scripts. The 40 languages shown in \textcolor{benchpara}{\textbf{green}}
have translated paraphrase variants available.}
\label{tab:language-list}

\footnotesize
\setlength{\tabcolsep}{5pt}
\renewcommand{\arraystretch}{1.10}

\definecolor{benchpara}{RGB}{34,139,34}

\newcommand{\benchLangSep}{\hspace{0.34em}\allowbreak}

\newcommand{\benchCode}[1]{\mbox{\texttt{#1}}\benchLangSep\ignorespaces}

\newcommand{\benchParaCode}[1]{\mbox{\textcolor{benchpara}{\bfseries\texttt{#1}}}\benchLangSep\ignorespaces}

\newcommand{\benchFull}[1]{\mbox{\texttt{\detokenize{#1}}}\benchLangSep\ignorespaces}

\newcommand{\benchParaFull}[1]{\mbox{\textcolor{benchpara}{\bfseries\texttt{\detokenize{#1}}}}\benchLangSep\ignorespaces}

\begin{tabularx}{\textwidth}{
  @{}
  >{\raggedright\arraybackslash}p{0.14\textwidth}
  >{\raggedright\arraybackslash}p{0.12\textwidth}
  >{\raggedleft\arraybackslash}p{0.06\textwidth}
  >{\raggedright\arraybackslash}X
  @{}
}
\toprule

\textbf{Script}
&
\textbf{Resource tier}
&
\textbf{Count}
&
\textbf{Languages}
\\

\midrule

\multirow{3}{*}{\shortstack[l]{\textbf{Latin}\\[-1pt]
\textcolor{black!60}{Latn $\cdot$ 121}}}
& High
& 19
&
\benchCode{cat} \benchCode{ces} \benchCode{dan}
\benchParaCode{deu} \benchCode{en} \benchCode{est}
\benchCode{fin} \benchParaCode{fra} \benchCode{hrv}
\benchCode{hun} \benchParaCode{ita} \benchCode{nld}
\benchCode{nob} \benchCode{pol} \benchParaCode{por}
\benchCode{slk} \benchParaCode{spa} \benchCode{swe}
\benchParaCode{vie}
\\

& Medium
& 25
&
\benchParaCode{afr} \benchCode{als} \benchParaCode{azj}
\benchCode{bos} \benchCode{ceb} \benchParaCode{cym}
\benchCode{epo} \benchParaCode{eus} \benchParaCode{gle}
\benchCode{glg} \benchParaCode{hau} \benchParaCode{ind}
\benchParaCode{isl} \benchParaCode{jav} \benchCode{lit}
\benchCode{lvs} \benchCode{mlt} \benchCode{nno}
\benchParaCode{ron} \benchCode{slv} \benchParaCode{swh}
\benchParaCode{tgl} \benchParaCode{tur} \benchParaCode{uzn}
\benchParaCode{zsm}
\\

& Low
& 77
&
\benchCode{ace} \benchCode{aka} \benchCode{arb}
\benchCode{ast} \benchCode{bam} \benchCode{ban}
\benchCode{bem} \benchCode{bjn} \benchCode{bug}
\benchCode{cjk} \benchCode{crh} \benchCode{dik}
\benchCode{dyu} \benchCode{ewe} \benchCode{fao}
\benchCode{fon} \benchCode{fur} \benchCode{fuv}
\benchCode{gaz} \benchCode{gla} \benchCode{grn}
\benchCode{hat} \benchCode{ibo} \benchCode{ilo}
\benchCode{kab} \benchCode{kac} \benchCode{kam}
\benchCode{kbp} \benchCode{kea} \benchCode{kin}
\benchCode{kmb} \benchCode{knc} \benchCode{kon}
\benchCode{lij} \benchCode{lim} \benchCode{lmo}
\benchCode{ltg} \benchCode{ltz} \benchCode{lua}
\benchCode{lug} \benchCode{luo} \benchCode{lus}
\benchParaCode{min} \benchCode{mos} \benchCode{mri}
\benchCode{nso} \benchCode{nus} \benchCode{nya}
\benchCode{oci} \benchCode{pag} \benchCode{pap}
\benchCode{plt} \benchCode{quy} \benchCode{run}
\benchCode{sag} \benchCode{scn} \benchCode{smo}
\benchCode{sna} \benchCode{som} \benchCode{sot}
\benchCode{srd} \benchCode{ssw} \benchCode{sun}
\benchCode{szl} \benchCode{taq} \benchCode{tpi}
\benchCode{tsn} \benchCode{tso} \benchCode{tuk}
\benchCode{tum} \benchCode{twi} \benchCode{umb}
\benchCode{vec} \benchCode{war} \benchCode{wol}
\benchCode{xho} \benchCode{zul}
\\

\midrule

\multirow{3}{*}{\shortstack[l]{\textbf{Arabic}\\[-1pt]
\textcolor{black!60}{Arab $\cdot$ 20}}}
& High
& 1
&
\benchParaCode{arb}
\\

& Medium
& 3
&
\benchParaCode{arz} \benchParaCode{pes} \benchParaCode{urd}
\\

& Low
& 16
&
\benchCode{ace} \benchCode{acm} \benchCode{acq}
\benchCode{aeb} \benchCode{ajp} \benchCode{apc}
\benchCode{ars} \benchCode{ary} \benchCode{azb}
\benchCode{bjn} \benchCode{kas} \benchCode{knc}
\benchCode{min} \benchCode{pbt} \benchCode{prs}
\benchCode{snd}
\\

\midrule

\multirow{3}{*}{\shortstack[l]{\textbf{Cyrillic}\\[-1pt]
\textcolor{black!60}{Cyrl $\cdot$ 11}}}
& High
& 2
&
\benchCode{bul} \benchParaCode{rus}
\\

& Medium
& 6
&
\benchParaCode{bel} \benchParaCode{kaz} \benchCode{khk}
\benchCode{mkd} \benchParaCode{srp} \benchParaCode{ukr}
\\

& Low
& 3
&
\benchCode{kir} \benchCode{tat} \benchCode{tgk}
\\

\midrule

\multirow{3}{*}{\shortstack[l]{\textbf{Devanagari}\\[-1pt]
\textcolor{black!60}{Deva $\cdot$ 7}}}
& High
& 1
&
\benchParaCode{hin}
\\

& Medium
& 1
&
\benchParaCode{mar}
\\

& Low
& 5
&
\benchCode{awa} \benchCode{bho} \benchCode{mag}
\benchCode{npi} \benchCode{san}
\\

\midrule

\multirow{3}{*}{\shortstack[l]{\textbf{Other scripts}\\[-1pt]
\textcolor{black!60}{14 scripts $\cdot$ 15}}}
& High
& 4
&
\benchParaFull{jpn_Jpan} \benchParaFull{kor_Hang}
\benchParaFull{zho_Hans} \benchFull{zho_Hant}
\\

& Medium
& 11
&
\benchParaFull{ben_Beng} \benchFull{guj_Gujr}
\benchFull{heb_Hebr} \benchFull{hye_Armn}
\benchFull{kan_Knda} \benchParaFull{kat_Geor}
\benchFull{mal_Mlym} \benchParaFull{tam_Taml}
\benchFull{tel_Telu} \benchParaFull{tha_Thai}
\benchFull{yue_Hant}
\\

& Low
& 0
&
\textcolor{black!45}{---}
\\

\midrule

\textbf{All scripts}
&
&
\textbf{174}
&
\textcolor{black!60}{27 high-resource, 46 medium-resource, and 101 low-resource entries}
\\

\bottomrule
\end{tabularx}
\end{threeparttable}
\end{table}

\subsection{Paraphrases}
\label{app:paraphrase}

We generate syntactically diverse paraphrases of every forget and retain question following the Extended Paraphrase Typology \citep{kovatchev-etal-2018-etpc}. \autoref{tab:ept_paraphrase_types} lists the 25 atomic types under their seven parent categories and defines each with an example; the twenty-fifth type, identity, leaves the wording unchanged and serves as the control.

Paraphrases are generated in English with GPT-5.4 from the original TOFU question and its gold answer. Over the 4,000 \texttt{forget10}+\texttt{retain90} questions this yields 52,916 applied question paraphrases, since not all paraphrase types were suitable for every data sample.

\paragraph{Translated paraphrases.}
English paraphrases are translated into the 40 languages marked in green in \autoref{tab:language-list}. This set includes 12 high resource languages, 27 medium resource languages, and one low resource language.

\begin{table}[htbp]
\centering
\caption{The 25 atomic paraphrase types in the ETPC \citep{kovatchev-etal-2018-etpc}. Examples are given
per category; the second line is a paraphrase of the first.}
\label{tab:ept_paraphrase_types}
\label{tab:ept-paraphrase-summary}

\scriptsize
\setlength{\tabcolsep}{4pt}
\renewcommand{\arraystretch}{1.05}

\newcommand{\eptcat}[1]{\textbf{#1}}

\newcommand{\eptexlab}{\dimexpr0.12\textwidth+2\tabcolsep\relax}
\newcommand{\eptexbody}{\dimexpr0.88\textwidth-2\tabcolsep\relax}
\newcommand{\eptex}[2]{\makebox[\eptexlab][l]{\textcolor{black!55}{\emph{Example}}}\begin{minipage}[t]{\eptexbody}#1\\[0.5pt]#2\end{minipage}}

\begin{tabularx}{\textwidth}{
    @{}
    >{\raggedright\arraybackslash}p{0.12\textwidth}
    >{\raggedright\arraybackslash}p{0.27\textwidth}
    >{\raggedright\arraybackslash}X
    @{}
}
\toprule

\textbf{Category}
&
\textbf{Paraphrase type}
&
\textbf{Definition}
\\

\midrule

\multirow[t]{3}{\linewidth}{\eptcat{Morphology}}
&
\textbf{Inflectional changes}
&
Changes grammatical inflection, such as number, tense, or person.
\\

&
\textbf{Modal verb changes}
&
Expresses the same proposition using a different modal construction.
\\

&
\textbf{Derivational changes}
&
Uses a morphologically related word.
\\

\addlinespace[3pt]
\multicolumn{3}{@{}p{\textwidth}@{}}{\eptex{How has Elvin Mammadov contributed to fiction literature?}{How has Elvin Mammadov contributed to literary fiction?}}
\\
\addlinespace[1.5pt]
\cmidrule(lr){1-3}
\addlinespace[1.5pt]

\multirow[t]{5}{\linewidth}{\eptcat{Lexicon}}
&
\textbf{Spelling changes}
&
Uses alternative spellings of the same word.
\\

&
\textbf{Same polarity substitution} {\color{black!55}(habitual)}
&
Replaces an expression with an applicable synonym.
\\

&
\textbf{Same polarity substitution} {\color{black!55}(contextual)}
&
Replaces an expression with one that has the same meaning.
\\

&
\textbf{Same polarity substitution} {\color{black!55}(named entity)}
&
Replaces a named entity with another expression referring to the same entity.
\\

&
\textbf{Change of format}
&
Represents the same information using a different textual format.
\\

\addlinespace[3pt]
\multicolumn{3}{@{}p{\textwidth}@{}}{\eptex{What is Rajeev Majumdar's birth date?}{What is Rajeev Majumdar's birthdate?}}
\\
\addlinespace[1.5pt]
\cmidrule(lr){1-3}
\addlinespace[1.5pt]

\multirow[t]{4}{\linewidth}{\eptcat{Lexico-syntax}}
&
\textbf{Opposite polarity substitution} {\color{black!55}(habitual)}
&
Uses a conventional opposite while restructuring the sentence.
\\

&
\textbf{Opposite polarity substitution} {\color{black!55}(contextual)}
&
Uses a context dependent opposite with structural changes.
\\

&
\textbf{Synthetic/analytic substitution}
&
Switches between a single lexical item and an equivalent multi word construction.
\\

&
\textbf{Converse substitution}
&
Uses the converse of a relation while swapping the roles of its arguments.
\\

\addlinespace[3pt]
\multicolumn{3}{@{}p{\textwidth}@{}}{\eptex{What professions do Hina Ameen's parents hold?}{What occupations do Hina Ameen's parents hold?}}
\\
\addlinespace[1.5pt]
\cmidrule(lr){1-3}
\addlinespace[1.5pt]

\multirow[t]{5}{\linewidth}{\eptcat{Syntax}}
&
\textbf{Diathesis alternation}
&
Changes how semantic roles map to syntax, such as active to passive voice.
\\

&
\textbf{Negation switching}
&
Changes the placement or expression of negation while preserving meaning.
\\

&
\textbf{Ellipsis}
&
Omits material that can be recovered from context.
\\

&
\textbf{Coordination changes}
&
Changes how phrases or clauses are joined through coordination.
\\

&
\textbf{Subordination and nesting changes}
&
Reorganizes information across main, subordinate or embedded clauses.
\\

\addlinespace[3pt]
\multicolumn{3}{@{}p{\textwidth}@{}}{\eptex{When did Behrouz Rohani publish his first Star Wars book?}{When was Behrouz Rohani's first Star Wars book published?}}
\\
\addlinespace[1.5pt]
\cmidrule(lr){1-3}
\addlinespace[1.5pt]

\multirow[t]{4}{\linewidth}{\eptcat{Discourse}}
&
\textbf{Punctuation changes}
&
Changes punctuation.
\\

&
\textbf{Direct/indirect style alternations}
&
Switches between direct quotation and reported speech.
\\

&
\textbf{Sentence modality changes}
&
Changes sentence form, such as declarative, interrogative, or imperative.
\\

&
\textbf{Syntax/discourse structure changes}
&
Reorganizes information across clauses or sentences.
\\

\addlinespace[3pt]
\multicolumn{3}{@{}p{\textwidth}@{}}{\eptex{What themes does Hina Ameen explore in her book, ``Shale Stories''?}{What themes does Hina Ameen explore in her book, Shale Stories?}}
\\
\addlinespace[1.5pt]
\cmidrule(lr){1-3}
\addlinespace[1.5pt]

\multirow[t]{3}{\linewidth}{\eptcat{Other}}
&
\textbf{Addition/Deletion}
&
Adds or removes information while preserving approximately the same overall meaning.
\\

&
\textbf{Change of order}
&
Reorders words, phrases, clauses, or information.
\\

&
\textbf{Semantic} {\color{black!55}(general inferences)}
&
Expresses equivalent information through semantic inference rather than surface transformation.
\\

\addlinespace[3pt]
\multicolumn{3}{@{}p{\textwidth}@{}}{\eptex{How has Elvin Mammadov's work impacted society and the literary world?}{How has Elvin Mammadov's work impacted the literary world and society?}}
\\
\addlinespace[1.5pt]
\cmidrule(lr){1-3}
\addlinespace[1.5pt]

\multirow[t]{1}{\linewidth}{\eptcat{Extremes}}
&
\textbf{Identity}
&
Leaves the wording unchanged; the control against which every other type is compared.
\\
\addlinespace[1.5pt]

\bottomrule
\end{tabularx}
\end{table}

\subsection{Checkpoint zoo}
\label{app:checkpoints}

Our results compare performance across a base model $M_{\mathrm{B}}$; a fine-tuned checkpoint $M_{\mathrm{FT}}$ trained on $D$; a retain-only oracle $M_{\mathrm{O}}$ trained on $D\setminus\Fset$; and unlearned endpoints $M'$ produced from $M_{\mathrm{FT}}$. We use three model families: Tiny Aya Global, Qwen3-4B-Instruct, and Llama-3.2-3B-Instruct.

\autoref{tab:model-checkpoint-coverage} records our checkpoint zoo: 103 fine-tuned checkpoints (81 monolingual, 22 multilingual), 103 matched oracles, and 287 unlearned checkpoints. The latter comprise 243 monolingual endpoints (three per fine-tuned checkpoint, one each for SimNPO, RMU, and GradDiff) and 44 Aya trilingual endpoints.
\begin{table}[htbp]
\centering
\begin{threeparttable}
\caption{Model and checkpoint coverage. Panel A summarizes the saved
checkpoints by model family. Panel B gives the multilingual language sets.}
\label{tab:model-checkpoint-coverage}

\scriptsize
\setlength{\tabcolsep}{5pt}
\renewcommand{\arraystretch}{1.10}

\newcommand{\benchFO}{\ensuremath{\mathrm{F}/\mathrm{O}}}

\newcommand{\benchFOU}[1]{\ensuremath{\mathrm{F}/\mathrm{O}/\mathrm{U}_{#1}}}

\begin{tabularx}{\textwidth}{
  @{}
  >{\raggedright\arraybackslash}p{0.16\textwidth}
  >{\raggedright\arraybackslash}X
  >{\centering\arraybackslash}p{0.115\textwidth}
  >{\centering\arraybackslash}p{0.145\textwidth}
  >{\centering\arraybackslash}p{0.16\textwidth}
  @{}
}
\toprule

\multicolumn{5}{@{}l}{\textbf{Checkpoint inventory}}
\\[2pt]

\textbf{Model family}
&
\textbf{Training regime}
&
\textbf{FT}
&
\textbf{Retain-90 oracle}
&
\textbf{Unlearned checkpoints}
\\

\midrule

\multirow{3}{*}{\textbf{Tiny Aya}}
& Monolingual
& 27 & 27 & 81
\\

& Trilingual endpoint cohort
& 7 & 7 & 44
\\

& Additional five-/six-language mixtures
& 5 & 5 & ---
\\

\addlinespace[1pt]

\multirow{2}{*}{\textbf{Qwen3}}
& Monolingual
& 27 & 27 & 81
\\

& Additional five-/six-language mixtures
& 5 & 5 & ---
\\

\addlinespace[1pt]

\multirow{2}{*}{\textbf{Llama}}
& Monolingual
& 27 & 27 & 81
\\

& Additional five-/six-language mixtures
& 5 & 5 & ---
\\

\midrule

\textbf{Total}
&
& \textbf{103}
& \textbf{103}
& \textbf{287}
\\

\bottomrule
\end{tabularx}

\vspace{7pt}

\begin{tabularx}{\textwidth}{
  @{}
  >{\raggedright\arraybackslash}X
  >{\centering\arraybackslash}p{0.055\textwidth}
  >{\centering\arraybackslash}p{0.19\textwidth}
  >{\centering\arraybackslash}p{0.17\textwidth}
  >{\centering\arraybackslash}p{0.17\textwidth}
  @{}
}
\toprule

\multicolumn{5}{@{}l}{\textbf{Multilingual training configurations}}
\\[2pt]

\textbf{Training mixture}
&
\shortstack{\textbf{Lang.}\\\textbf{count}}
&
\textbf{Tiny Aya}
&
\textbf{Qwen3}
&
\textbf{Llama}
\\

\midrule

\multicolumn{5}{@{}l}{\textcolor{black!60}{\textit{Three-language mixtures}}}
\\[-1pt]

\texttt{DE + EN + NL}
& 3 & \benchFOU{6} & --- & ---
\\

\texttt{DE + EN + ES}
& 3 & \benchFOU{9} & --- & ---
\\

\texttt{EN + RU + ZH}
& 3 & \benchFOU{9} & --- & ---
\\

\texttt{BE + RU + UK}
& 3 & \benchFOU{4} & --- & ---
\\

\texttt{IT + PT + ES}
& 3 & \benchFOU{4} & --- & ---
\\

\texttt{EN + ES + ZH}
& 3 & \benchFOU{9} & --- & ---
\\

\texttt{EN + ID + TR}
& 3 & \benchFOU{3} & --- & ---
\\

\addlinespace[2pt]
\cmidrule(lr){1-5}
\addlinespace[2pt]

\multicolumn{5}{@{}l}{\textcolor{black!60}{\textit{Five- and six-language mixtures}}}
\\[-1pt]

\texttt{AR + EN + JA + RU + ES + ZH}
& 6 & \benchFO & \benchFO & \benchFO
\\

\texttt{DE + EN + FR + IT + ES}
& 5 & \benchFO & \benchFO & \benchFO
\\

\texttt{DE + EN + ID + SW + TR + VI}
& 6 & \benchFO & \benchFO & \benchFO
\\

\texttt{EN + FR + IT + PT + ES}
& 5 & \benchFO & \benchFO & \benchFO
\\

\texttt{EN + AR + HA + HI + NE + SW}
& 6 & \benchFO & \benchFO & \benchFO
\\

\midrule

\textbf{Available multilingual F/O pairs}
&
& \textbf{12}
& \textbf{5}
& \textbf{5}
\\

\bottomrule
\end{tabularx}

\begin{tablenotes}[flushleft]
\tiny

\item
Each monolingual fine-tuned checkpoint has three unlearned checkpoints,
one each for SimNPO, RMU, and GradDiff. Trilingual counts include all saved
method-specific endpoints.

\item
In Panel B, \ensuremath{\mathrm{F}} denotes a fine-tuned checkpoint,
\ensuremath{\mathrm{O}} a retain-90 oracle, and
\ensuremath{\mathrm{U}_{n}} $n$ saved unlearned checkpoints.

\end{tablenotes}
\end{threeparttable}
\end{table}

\clearpage
\subsection{Model choice and multilingual capability}
\label{app:model-capability}
The purpose of this appendix is to justify the choice of Qwen3-4B-Instruct-2507 \citep{yang2025qwen3technicalreport}, Llama-3.2-3B-Instruct \citep{grattafiori2024llama3herdmodels} and Tiny Aya Global \citep{salamanca2026tinyayabridgingscale}, as our multilingual LMs. Qwen3-4B reports proficiency in over 100 languages, while Tiny Aya Global is explicitly trained for stronger performance on many low resource languages. Llama-3.2-3B serves as a third control. We do not expect any model to be proficient in all 174 languages, but given that any user could query  these models in any language, we see value in our design to evaluate in all languages.

We measure these models' capabilities on Belebele \citep{bandarkar-etal-2024-belebele} and Global MMLU \citep{singh-etal-2025-global}. \autoref{tab:model-capability-belebele} shows the Belebele results for the 46 evaluated languages. \autoref{tab:acquisition-parent-capability} shows each model family after fine-tuning in our language sets (script diverse, Latin only and low resource). The results demonstrate sufficient retention and answer capabilities across all languages used in our source selection experiments.

\begin{table}[tbp]
\centering
\begin{threeparttable}
\caption{Capability of the language-set specific model families before unlearning. All entries are percentages.}
\label{tab:acquisition-parent-capability}

\footnotesize
\setlength{\tabcolsep}{3.5pt}
\renewcommand{\arraystretch}{1.08}

\begin{tabularx}{\textwidth}{@{}
    >{\RaggedRight\arraybackslash}p{0.18\textwidth}
    >{\RaggedRight\arraybackslash}X
    >{\centering\arraybackslash}p{0.12\textwidth}
    >{\centering\arraybackslash}p{0.15\textwidth}
    >{\centering\arraybackslash}p{0.12\textwidth}
    @{}}
\toprule
Inventory &
Acquisition parent &
\makecell[c]{Mean full\\recovery} &
\makecell[c]{Lowest full\\recovery} &
\makecell[c]{Mean\\target-only} \\
\midrule

\multirow{3}{*}{\makecell[l]{Original /\\script-diverse}}
  & Tiny Aya Global              & 88.56 & 76.00 (JA)     & 99.39 \\
  & Qwen3-4B-Instruct-2507       & 91.17 & 76.33 (JA)     & 99.39 \\
  & Llama-3.2-3B-Instruct        & 93.61 & 90.67 (AR)     & 99.33 \\

\midrule
\multirow{3}{*}{Latin-script}
  & Tiny Aya Global              & 92.33 & 86.00 (TR)     & 99.67 \\
  & Qwen3-4B-Instruct-2507       & 94.33 & 92.00 (DE, ID) & 98.67 \\
  & Llama-3.2-3B-Instruct        & 97.00 & 92.00 (SW)     & 99.67 \\

\midrule
\multirow{3}{*}{Lower-resource}
  & Tiny Aya Global              & 88.50 & 77.98 (NE)     & 99.31 \\
  & Qwen3-4B-Instruct-2507       & 81.84 & 75.44 (NE)     & 99.12 \\
  & Llama-3.2-3B-Instruct        & 88.83 & 72.28 (NE)     & 99.13 \\

\bottomrule
\end{tabularx}
\end{threeparttable}
\end{table}

\clearpage
\begin{table}[tbp]
\centering
\caption{Released checkpoint Belebele accuracy across the 46 evaluated
languages.}
\label{tab:model-capability-belebele}

\begingroup

\scriptsize
\setlength{\tabcolsep}{2.7pt}
\renewcommand{\arraystretch}{1.03}

\noindent
\begin{minipage}[t]{0.485\textwidth}
\vspace{0pt}
\begin{tabularx}{\linewidth}{@{}
    >{\RaggedRight\arraybackslash}X r r r @{}}
\toprule
Language &
\makecell[c]{Tiny Aya\\Global} &
\makecell[c]{Qwen3\\4B} &
\makecell[c]{Llama 3.2\\3B} \\
\midrule
Amharic                  & 48.6 & 36.1 & 26.8 \\
Arabic         & 63.0 & 85.4 & 58.3 \\
Bengali                  & 59.2 & 72.0 & 44.0 \\
Czech                    & 68.7 & 84.3 & 55.9 \\
German         & 67.8 & 88.7 & 51.4 \\
Greek                    & 68.2 & 82.6 & 43.4 \\
English        & 74.4 & 91.4 & 83.6 \\
Spanish        & 67.2 & 86.7 & 57.3 \\
Persian                  & 62.6 & 78.3 & 58.4 \\
Filipino/Tagalog         & 66.6 & 76.0 & 45.9 \\
French                   & 70.6 & 88.9 & 61.8 \\
Hausa          & 50.1 & 28.1 & 30.1 \\
Hebrew                   & 64.1 & 77.8 & 33.0 \\
Hindi          & 57.9 & 71.8 & 41.8 \\
Indonesian     & 69.3 & 85.9 & 45.8 \\
Igbo                     & 40.1 & 27.4 & 25.7 \\
Italian                  & 68.6 & 86.2 & 37.1 \\
Japanese       & 61.6 & 81.7 & 61.3 \\
Korean                   & 64.9 & 84.1 & 60.1 \\
Kyrgyz                   & 40.8 & 58.6 & 34.1 \\
Lithuanian               & 69.1 & 78.8 & 40.4 \\
Malagasy                 & 53.7 & 35.4 & 25.7 \\
Malay                    & 69.0 & 83.3 & 50.9 \\
\bottomrule
\end{tabularx}
\end{minipage}
\hfill
\begin{minipage}[t]{0.485\textwidth}
\vspace{0pt}
\begin{tabularx}{\linewidth}{@{}
    >{\RaggedRight\arraybackslash}X r r r @{}}
\toprule
Language &
\makecell[c]{Tiny Aya\\Global} &
\makecell[c]{Qwen3\\4B} &
\makecell[c]{Llama 3.2\\3B} \\
\midrule
Nepali         & 56.7 & 64.9 & 32.4 \\
Dutch                    & 67.9 & 86.0 & 65.9 \\
Nyanja/Chichewa          & 29.0 & 29.7 & 25.6 \\
Polish                   & 67.3 & 83.4 & 48.1 \\
Portuguese               & 69.0 & 81.8 & 22.9 \\
Romanian                 & 69.1 & 85.7 & 45.4 \\
Russian        & 71.2 & 85.0 & 70.1 \\
Sinhala                  & 30.6 & 56.3 & 27.3 \\
Shona                    & 49.6 & 33.9 & 27.0 \\
Somali                   & 30.3 & 29.9 & 25.2 \\
Serbian                  & 67.4 & 81.8 & 46.4 \\
Swedish                  & 69.9 & 86.2 & 61.2 \\
Swahili        & 61.8 & 43.8 & 36.9 \\
Telugu                   & 54.1 & 63.6 & 37.2 \\
Turkish        & 63.7 & 79.6 & 49.1 \\
Ukrainian                & 67.9 & 82.2 & 58.4 \\
Vietnamese     & 67.4 & 86.0 & 56.4 \\
Yoruba                   & 39.9 & 30.4 & 26.3 \\
Chinese        & 67.9 & 90.0 & 70.7 \\
Burmese/Myanmar          & 52.9 & 53.3 & 29.0 \\
Tamil                    & 59.3 & 67.2 & 41.4 \\
Thai                     & 61.9 & 81.3 & 54.4 \\
Urdu                     & 58.2 & 71.1 & 48.7 \\
\bottomrule
\end{tabularx}
\end{minipage}
\endgroup
\end{table}

\FloatBarrier

\subsection{Generation judge and extraction strength}
\label{app:generation-judge}
Open ended generations are scored with an LLM as judge for semantic leakage and generation failures. The implementation extracts the question, gold answer, model generation, evaluation language, and expected answer language. We use \qwenj as the judge with temperature zero. Each example is prompted with the question, gold answer, model generation, expected answer language, the failure definitions, decision rules, and few shot examples.
Each gold answer is decomposed into atomic facts, and the judge grades every atomic fact of a generation independently as \texttt{none}, \texttt{partial} or \texttt{full} disclosure. The judge separately flags a generation as \emph{gibberish} (unreadable or degenerate output) and as \emph{code-switching} (an answer not in the requested language); the two flags can co-occur, and disclosure is counted regardless of them. \emph{Full leakage} is the share of forget generations whose atomic facts are all judged \texttt{full}; \emph{any disclosure} is the share with at least one fact judged \texttt{partial} or \texttt{full}; \emph{retained recovery} applies the full criterion to retain facts, so higher is better. \emph{Extraction strength} (ES) measures the fraction of the correct answer that is recoverable as an exact suffix under teacher forcing \citep{dorna2025openunlearningacceleratingllmunlearning}.
\paragraph{Human calibration of the judge.}
We ran a human-LLM-judge three way calibration study. The human-LLM portion spanned 192 TOFU forget set facts. We employed a secondary LLM judge as well to measure agreement with \qwenj covering 11,040 judgments. The generations reviewed were produced by Llama-3.2-1B-Instruct checkpoints, labeled by \qwenj and our second judge (\gptj). We sampled 60 answers on which the two judges assign the same question-level label and 60 on which they differ, which is 8.8\% of the set. \qwenj matches the human on 81.8\% of fact statuses (Cohen's $\kappa = 0.64$). Agreement is highest for full leakage ($\kappa = 0.86$), where the judge's rate is 4.1 points below the human's ($[-9.9, +0.4]$).
\begin{table}[t]
\centering
\small
\caption{Agreement between \qwenj and a blind human rater on 192 facts. Brackets: 95\% stratified bootstrap intervals. Rate difference is the judge's rate minus the human's.}
\label{tab:judge-human-agreement}
\begin{tabular}{lccc}
\toprule
Label & Agree (\%) & Cohen's $\kappa$ & Rate difference (pts) \\
\midrule
Fact status (\texttt{none}/\texttt{partial}/\texttt{full}) & 81.8 & 0.64 [0.50, 0.77] & -- \\
Full leakage   & 94.7 & 0.86 [0.69, 0.98] & $-4.1$ [$-9.9$, $+0.4$] \\
Any disclosure & 88.6 & 0.77 [0.64, 0.87] & $-11.4$ [$-17.9$, $-6.2$] \\
\bottomrule
\end{tabular}
\end{table}

\clearpage
\section{Translation generation and quality screening}
\label{app:translation}

LLMs are increasingly used for translation \citep{kocmi-etal-2024-findings}. This carries risks for fidelity and explainability, with hallucinations and errors remaining a concern, especially in low resource settings \citep{guerreiro-etal-2023-hallucinations}. Nonetheless LLMs fill a gap that neural machine translation leaves for low-resource languages. We include in this appendix details on the choice of our translation engine and evaluations.
\subsection{Translation model selection}
\label{app:translation-engine}

We evaluated nine frontier models: DeepSeek-V3, GPT-4o, GPT-4.1, GPT-5-mini, GPT-5-chat, GPT-5, GPT-5.1, GPT-5.2-chat, and GPT-5.4-nano. For each English TOFU QA pair, the candidate translation and the NLLB-200 translation \citep{nllbteam2022languageleftbehindscaling} were passed to a blinded pairwise judge (GPT-5.4), which selected its preferred translation. \autoref{tab:translation_win_rates} reports the mean win rate against NLLB and the number of languages judged for each candidate on our 40-language set. We selected GPT-5.2-chat for its highest win rate and full language coverage. \autoref{tab:gpt52_nllb_judge_delta} shows performance across four judge criteria: meaning preservation, answer consistency, fluency, and preservation of names and numbers, each scored 1--5 against the English gold.

\begin{table}[!htbp]
\centering
\small
\begin{threeparttable}
\caption{
For each language and translation model, win rate is the fraction of items where the model translation won relative to NLLB.
GPT-5.2-chat had the highest reported mean while also providing full
language coverage.
}
\label{tab:translation_win_rates}
\begin{tabular}{lcc}
\toprule
\textbf{Translation model} &
\textbf{Languages judged} &
\textbf{Mean win rate vs.\ NLLB (\%)} $\uparrow$ \\
\midrule
\textbf{GPT-5.2-chat} & \textbf{40} & \textbf{94.8\%} \\
GPT-5        & 30 & 94.4\% \\
DeepSeek-V3  & 35 & 88.9\% \\
GPT-4.1      & 40 & 88.3\% \\
GPT-5.1      & 40 & 88.0\% \\
GPT-5-mini   & 35 & 87.2\% \\
GPT-4o       & 40 & 87.1\% \\
GPT-5.4-nano & 40 & 84.9\% \\
GPT-5-chat   & 40 & 82.4\% \\
\bottomrule
\end{tabular}

\begin{tablenotes}[flushleft]
\footnotesize
\item The 40 languages used for translation evaluation are listed in \autoref{tab:gpt52_nllb_judge_delta}.
\end{tablenotes}
\end{threeparttable}
\end{table}

\begin{table}[htbp]
\centering
\footnotesize
\setlength{\tabcolsep}{3.5pt}
\renewcommand{\arraystretch}{1.08}
\providecommand{\heatcell}{}
\definecolor{heatcolor}{RGB}{31,109,160}
\renewcommand{\heatcell}[1]{\ifdim #1pt<0.75pt \cellcolor{heatcolor!7}\else
  \ifdim #1pt<1.25pt \cellcolor{heatcolor!16}\else
  \ifdim #1pt<1.75pt \cellcolor{heatcolor!26}\else
  \ifdim #1pt<2.25pt \cellcolor{heatcolor!36}\else
  \ifdim #1pt<2.75pt \cellcolor{heatcolor!46}\else
  \ifdim #1pt<3.25pt \cellcolor{heatcolor!56}\else
                     \cellcolor{heatcolor!66}\fi\fi\fi\fi\fi\fi
  #1}
\caption{\textbf{GPT-5.2-chat translations beat NLLB on every criterion in every judged language.} Judge grades (1--5, against the English gold) for meaning, answer consistency, fluency and names/numbers; each cell is the GPT-5.2-chat grade minus the NLLB grade, so positive favors GPT-5.2-chat. Languages are sorted by the mean of the four differences.}
\label{tab:gpt52_nllb_judge_delta}
\begin{minipage}[t]{0.49\textwidth}\centering
\begin{tabular}{@{}lcccc@{}}
\toprule
\textbf{Language} & \textbf{Mean.} & \makecell{\textbf{Answer}\\\textbf{cons.}} & \textbf{Flu.} & \makecell{\textbf{Names/}\\\textbf{num.}} \\
\midrule
zho\_Hant & \heatcell{3.44} & \heatcell{3.25} & \heatcell{3.76} & \heatcell{2.17} \\
zho\_Hans & \heatcell{2.98} & \heatcell{2.67} & \heatcell{3.52} & \heatcell{1.84} \\
ukr\_Cyrl & \heatcell{2.45} & \heatcell{1.96} & \heatcell{3.21} & \heatcell{2.12} \\
heb\_Hebr & \heatcell{2.58} & \heatcell{2.20} & \heatcell{3.20} & \heatcell{1.76} \\
arb\_Arab & \heatcell{2.27} & \heatcell{1.82} & \heatcell{3.03} & \heatcell{1.83} \\
jpn\_Jpan & \heatcell{2.38} & \heatcell{1.85} & \heatcell{3.31} & \heatcell{1.37} \\
hye\_Armn & \heatcell{2.21} & \heatcell{1.70} & \heatcell{2.96} & \heatcell{1.71} \\
bel\_Cyrl & \heatcell{2.17} & \heatcell{1.58} & \heatcell{2.78} & \heatcell{1.92} \\
rus\_Cyrl & \heatcell{2.07} & \heatcell{1.46} & \heatcell{2.92} & \heatcell{1.97} \\
kat\_Geor & \heatcell{2.29} & \heatcell{1.79} & \heatcell{2.80} & \heatcell{1.22} \\
quy\_Latn & \heatcell{2.34} & \heatcell{2.50} & \heatcell{2.08} & \heatcell{0.83} \\
kor\_Hang & \heatcell{2.21} & \heatcell{1.63} & \heatcell{2.77} & \heatcell{1.02} \\
guj\_Gujr & \heatcell{2.07} & \heatcell{1.67} & \heatcell{2.76} & \heatcell{1.06} \\
uig\_Arab & \heatcell{2.19} & \heatcell{1.92} & \heatcell{2.31} & \heatcell{1.11} \\
ell\_Grek & \heatcell{2.09} & \heatcell{1.49} & \heatcell{2.71} & \heatcell{1.11} \\
khm\_Khmr & \heatcell{2.34} & \heatcell{1.88} & \heatcell{2.45} & \heatcell{0.67} \\
tur\_Latn & \heatcell{2.00} & \heatcell{1.56} & \heatcell{2.76} & \heatcell{0.97} \\
tha\_Thai & \heatcell{2.09} & \heatcell{1.59} & \heatcell{2.77} & \heatcell{0.84} \\
tel\_Telu & \heatcell{2.03} & \heatcell{1.64} & \heatcell{2.66} & \heatcell{0.94} \\
tam\_Taml & \heatcell{1.99} & \heatcell{1.62} & \heatcell{2.56} & \heatcell{1.04} \\
\midrule
 & & & & \\
\bottomrule
\end{tabular}
\end{minipage}\hfill
\begin{minipage}[t]{0.49\textwidth}\centering
\begin{tabular}{@{}lcccc@{}}
\toprule
\textbf{Language} & \textbf{Mean.} & \makecell{\textbf{Answer}\\\textbf{cons.}} & \textbf{Flu.} & \makecell{\textbf{Names/}\\\textbf{num.}} \\
\midrule
wol\_Latn & \heatcell{2.25} & \heatcell{2.31} & \heatcell{2.00} & \heatcell{0.61} \\
pes\_Arab & \heatcell{1.85} & \heatcell{1.39} & \heatcell{2.69} & \heatcell{1.23} \\
ben\_Beng & \heatcell{1.93} & \heatcell{1.35} & \heatcell{2.79} & \heatcell{0.95} \\
sin\_Sinh & \heatcell{1.92} & \heatcell{1.55} & \heatcell{2.43} & \heatcell{0.90} \\
urd\_Arab & \heatcell{1.81} & \heatcell{1.34} & \heatcell{2.71} & \heatcell{0.90} \\
amh\_Ethi & \heatcell{1.68} & \heatcell{1.49} & \heatcell{1.92} & \heatcell{1.28} \\
npi\_Deva & \heatcell{1.73} & \heatcell{1.23} & \heatcell{2.47} & \heatcell{0.89} \\
mya\_Mymr & \heatcell{1.93} & \heatcell{1.50} & \heatcell{1.99} & \heatcell{0.88} \\
deu\_Latn & \heatcell{1.70} & \heatcell{1.19} & \heatcell{2.45} & \heatcell{0.82} \\
yor\_Latn & \heatcell{2.02} & \heatcell{1.75} & \heatcell{1.80} & \heatcell{0.48} \\
dzo\_Tibt & \heatcell{1.75} & \heatcell{1.64} & \heatcell{1.59} & \heatcell{0.87} \\
fra\_Latn & \heatcell{1.53} & \heatcell{0.95} & \heatcell{2.38} & \heatcell{0.87} \\
spa\_Latn & \heatcell{1.49} & \heatcell{0.98} & \heatcell{2.12} & \heatcell{0.78} \\
por\_Latn & \heatcell{1.39} & \heatcell{0.91} & \heatcell{2.19} & \heatcell{0.80} \\
lao\_Laoo & \heatcell{1.66} & \heatcell{1.21} & \heatcell{1.89} & \heatcell{0.38} \\
hau\_Latn & \heatcell{1.51} & \heatcell{1.13} & \heatcell{1.76} & \heatcell{0.40} \\
grn\_Latn & \heatcell{1.54} & \heatcell{1.49} & \heatcell{1.25} & \heatcell{0.35} \\
ind\_Latn & \heatcell{1.24} & \heatcell{0.68} & \heatcell{1.91} & \heatcell{0.38} \\
zsm\_Latn & \heatcell{1.20} & \heatcell{0.68} & \heatcell{1.96} & \heatcell{0.31} \\
ibo\_Latn & \heatcell{1.20} & \heatcell{0.92} & \heatcell{1.13} & \heatcell{0.18} \\
\midrule
\textbf{Mean (40)} & \heatcell{1.99} & \heatcell{1.59} & \heatcell{2.47} & \heatcell{1.04} \\
\bottomrule
\end{tabular}
\end{minipage}
\end{table}

\subsection{Translation prompt}
\label{app:translation-prompt}

\begin{figure}[htbp]
\centering
\begin{minipage}{0.95\textwidth}
\scriptsize
\textbf{System message}
\begin{verbatim}
You are a precise dataset translator. Translate question-answer records from
English into the requested target language while preserving meaning and all
placeholders. Return only the JSON object required by the schema.
\end{verbatim}
\textbf{User message}
\begin{verbatim}
Target language: <LANGUAGE NAME> (<FLORES-200 CODE>)

Translate each record's question and answer into the target language.

Rules:
- Keep every placeholder token exactly unchanged, for example
  ZXQPROTECTED000ZXQ.
- Do not translate, modify, drop, or reorder placeholder tokens.
- Translate all non-placeholder English text into <LANGUAGE NAME>; do not
  leave English clauses or boilerplate untranslated.
- Do not change the answer choice/fact relationship.
- Do not add commentary.
- Return one translation object for every input id.

Input records JSON:
[{"id": "<record id>",
  "question": "<English question>",
  "answer": "<English answer>"},
 ... one entry per record in the batch ...]
\end{verbatim}
\end{minipage}
\caption{Translation prompt. \texttt{<LANGUAGE NAME>} and \texttt{<FLORES-200 CODE>} are substituted per language, e.g.\ \emph{Modern Standard Arabic} and \texttt{arb\_Arab}.}
\label{fig:translation-prompt}
\end{figure}

\subsection{Quality evaluation}
\label{app:translation-quality}

The established win rate against NLLB says nothing about absolute quality. We face a challenge with assessing our full corpus since evaluating translation quality in under resourced languages is hard given the scarcity of high quality evaluation data \citep{wang-etal-2024-afrimte}.

\begin{figure}[htbp]
\centering
\includegraphics[width=0.85\textwidth]{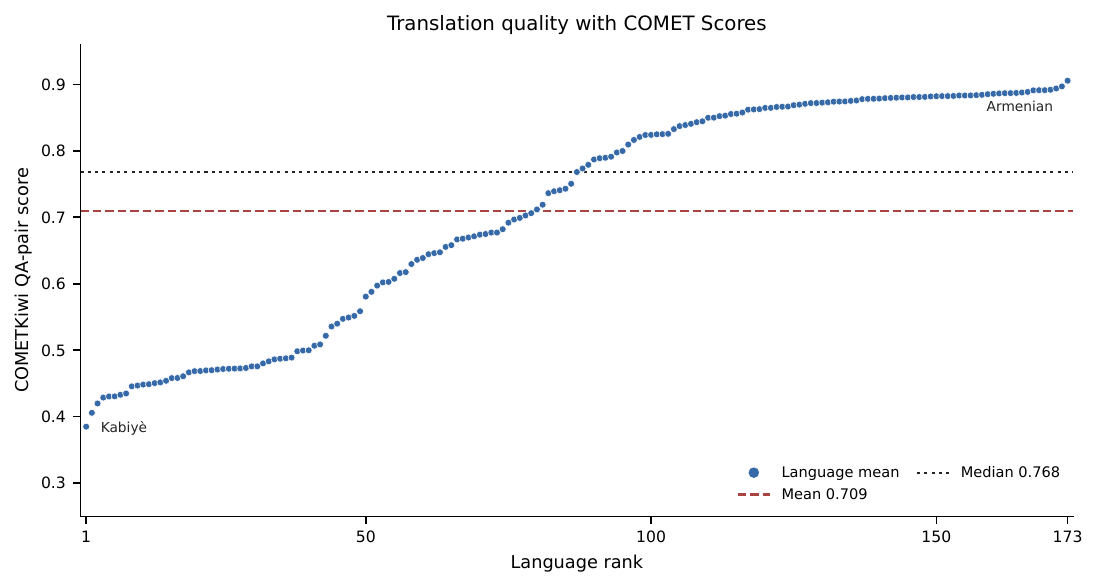}
\caption{Reference-free translation quality. Each point is one language's mean COMETKiwi score over the sampled QA pairs, a pair score averaging its question and answer, sorted from lowest to highest. The panel mean is 0.709 and the median 0.768}
\label{fig:translation-cometkiwi}
\end{figure}

\begin{figure}[htbp]
\centering
\includegraphics[width=0.9\textwidth]{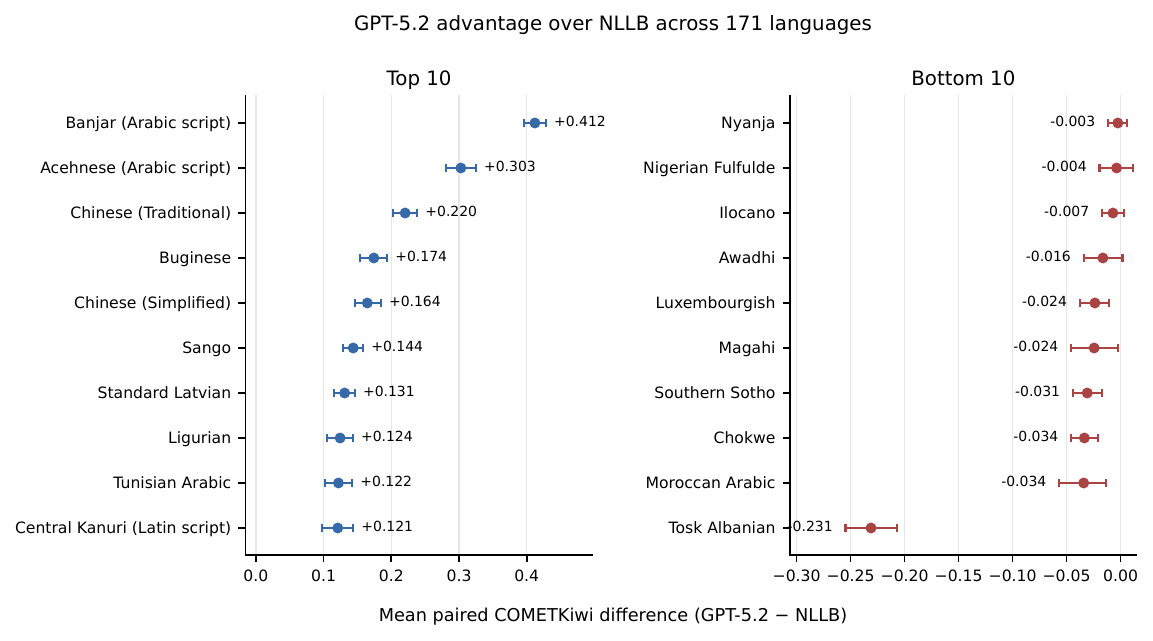}
\caption{Mean paired COMETKiwi difference (our translation minus NLLB), in COMETKiwi score units, over 200 aligned segments per language, for the ten languages with the largest advantage  and the ten with the smallest. Positive values favor our translation; negative values favor NLLB.}
\label{fig:translation-paired-comet}
\end{figure}

\begin{figure}[htbp]
\centering
\includegraphics[width=0.9\textwidth]{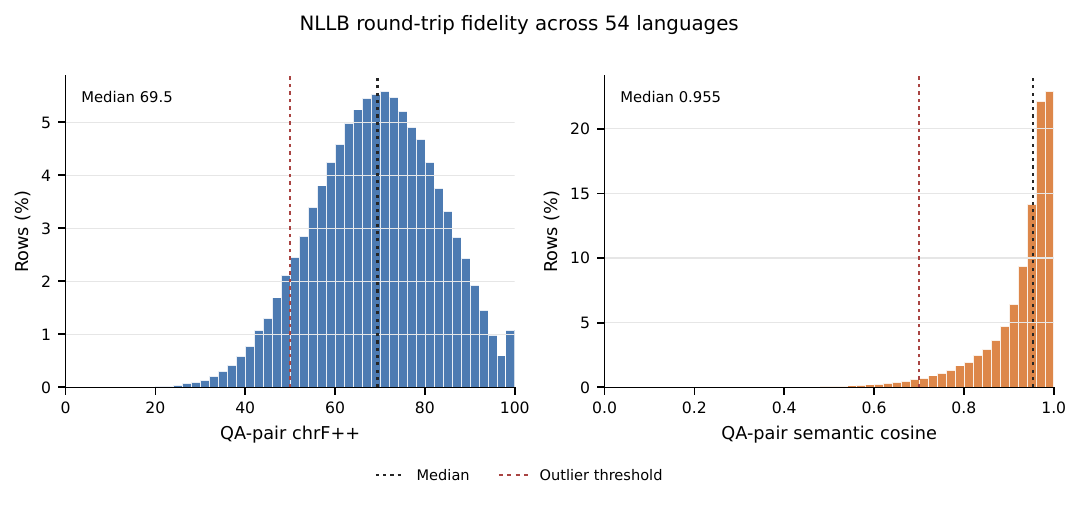}
\caption{NLLB round-trip fidelity over 54 languages. Distributions over QA pairs of chrF++ between the original English and the NLLB back-translation (left, median 69.5) and of embedding cosine similarity between the same two texts (right, median 0.955). Dotted lines mark the median and dashed lines the outlier threshold of the screen in Appendix~\ref{app:translation-screen}.}
\label{fig:translation-roundtrip}
\end{figure}

We therefore combine four signals:
\begin{enumerate}
    \item \textbf{Reference-free quality estimation.} COMETKiwi \citep{rei-etal-2022-cometkiwi} on the translations (\autoref{fig:translation-cometkiwi}).
    \item \textbf{Paired quality estimation.} COMETKiwi on our translation minus COMETKiwi on the NLLB translation (\autoref{fig:translation-paired-comet}).
    \item \textbf{Round-trip fidelity.} Back-translation to English with NLLB, compared with the original by chrF++ and embedding cosine similarity (\autoref{fig:translation-roundtrip}).
    \item \textbf{Pair-failure rate.} The fraction of QA pairs for which translation failed or returned a malformed record.
\end{enumerate}

\subsection{Language screening}
\label{app:translation-screen}

We compare each language against the cross-language distribution of each metric. A score is an extreme negative outlier when it falls more than three scaled median absolute deviations ($1.4826\,\mathrm{MAD}$) below the cross-language median, or above it for the pair-failure rate. A language is pruned when it is an outlier on two or more metrics. \autoref{tab:screening_thresholds} lists the cutoff for each metric and how many languages it flags; \autoref{tab:trust_classification} names the flagged languages. Four languages were pruned. ``Pruned'' means not eligible for selection for fine tuning.

\begin{table}[!htbp]
\centering
\small
\caption{
Screening rules and the number of languages flagged by each. Cutoffs for the
four primary metrics are set at the median minus
$3\times1.4826\times\mathrm{MAD}$; the pair-failure screen uses the fixed
criterion that more than half of QA pairs fail.
}
\label{tab:screening_thresholds}
\begin{tabular}{l r r}
\toprule
\textbf{Metric} &
\textbf{Screening rule} &
\textbf{Languages flagged} \\
\midrule
Direct COMETKiwi                     & Score $< 0.300964$  & 0  \\
GPT-5.2 $-$ NLLB COMETKiwi           & Score $< -0.062456$ & 1  \\
Back-translation chrF++              & Score $< 42.834452$ & 4  \\
Back-translation semantic similarity & Score $< 0.789314$  & 7  \\
Back-translation pair-failure rate   & Rate $> 50\%$       & 19 \\
\bottomrule
\end{tabular}
\end{table}

\begin{table}[!htbp]
\centering
\small
\setlength{\tabcolsep}{4pt}
\caption{
Pruned languages are almost exclusively low resource but only a handful were ultimately ruled out. Outlier metrics are the
four direct and back-translation metrics in \autoref{tab:screening_thresholds}.
}
\label{tab:trust_classification}
\begin{tabularx}{\textwidth}{
  @{}
  >{\RaggedRight\arraybackslash}p{0.31\textwidth}
  >{\raggedleft\arraybackslash}p{0.12\textwidth}
  Y
  @{}
}
\toprule
\textbf{Classification} &
\textbf{Languages} &
\textbf{Language names} \\
\midrule

Outlier on two or more metrics
& 4
& Chokwe, Dyula, Fon, Tosk Albanian \\
\addlinespace

Outlier on one metric
& 3
& Central Kanuri (Arabic), Jingpho, Southwestern Dinka \\
\addlinespace

NLLB diagnostics unavailable
& 2
& Minangkabau (Arabic), Modern Standard Arabic (Romanized) \\
\addlinespace

No issues
& 152
& See supplementary table \\

\bottomrule
\end{tabularx}
\end{table}

\clearpage
\section{Benchmark findings}
\label{app:findings}
This appendix gives the full tables behind the benchmark findings in Section~\ref{sec:benchmark}: how acquisition and unlearning transfer across languages, how the two agree, and how the form of the question changes what remains accessible.

\FloatBarrier
\subsection{Acquisition transfer and language rankings}
\label{app:findings-transfer-summary}
\label{app:findings-acquisition-rankings}
\begin{table}[htbp]
\centering
\caption{\textbf{Acquisition transfer rankings.} Mean fine-tuned-minus-base gold-answer likelihood gain per model family (common 21-language acquisition cohort, 50 forget facts per cell; no oracle subtraction or eligibility filter).
(a) Top ten acquisition (source) languages by mean gain over the 173 other language--script pairs, with question and answer in the target language.
(b) Top ten receiving question languages, with answers kept in the acquisition language.
(c) Top ten receiving languages, with both question and answer in the named language.}
\label{tab:acquisition-rankings}
\label{tab:acquisition-donors-top10}
\label{tab:acquisition-receivers-training-answers-top10}
\label{tab:acquisition-receivers-native-answers-top10}
\begingroup\small
\setlength{\tabcolsep}{2pt}\renewcommand{\arraystretch}{1.08}
\begin{tabular*}{\textwidth}{@{\extracolsep{\fill}}r lr lr lr@{}}
\toprule
\multicolumn{7}{@{}l}{\textbf{(a) Source languages: gain over 173 other language--script pairs}} \\
\addlinespace[2pt]
 & \multicolumn{2}{c}{Aya} & \multicolumn{2}{c}{Qwen} & \multicolumn{2}{c}{Llama} \\
\cmidrule(lr){2-3}\cmidrule(lr){4-5}\cmidrule(lr){6-7}
Rank & Language & Gain & Language & Gain & Language & Gain \\
\midrule
1 & Spanish & 0.1664 & Spanish & 0.1335 & Spanish & 0.0667 \\
2 & Norwegian Bokm{\aa}l & 0.1621 & Vietnamese & 0.1264 & German & 0.0625 \\
3 & Indonesian & 0.1616 & Norwegian Bokm{\aa}l & 0.1262 & Vietnamese & 0.0596 \\
4 & Vietnamese & 0.1558 & Indonesian & 0.1259 & Thai & 0.0531 \\
5 & Thai & 0.1557 & German & 0.1235 & Indonesian & 0.0525 \\
6 & English & 0.1537 & Thai & 0.1179 & Turkish & 0.0493 \\
7 & German & 0.1508 & Korean & 0.1110 & Korean & 0.0480 \\
8 & Hausa & 0.1482 & Turkish & 0.1104 & Japanese & 0.0447 \\
9 & Turkish & 0.1438 & English & 0.1098 & Norwegian Bokm{\aa}l & 0.0442 \\
10 & Zulu & 0.1437 & Tamil & 0.1073 & English & 0.0412 \\
\midrule
\addlinespace[3pt]
\multicolumn{7}{@{}l}{\textbf{(b) Receiving question languages, answers in the acquisition language}} \\
\addlinespace[2pt]
Rank & Language & Gain & Language & Gain & Language & Gain \\
\midrule
1 & English & +0.6594 & English & +0.6415 & English & +0.6364 \\
2 & French & +0.6435 & Spanish & +0.6162 & Spanish & +0.5824 \\
3 & Portuguese & +0.6360 & Traditional Chinese & +0.6136 & Portuguese & +0.5719 \\
4 & Italian & +0.6353 & Portuguese & +0.6128 & Asturian & +0.5640 \\
5 & Spanish & +0.6343 & French & +0.6009 & French & +0.5576 \\
6 & Indonesian & +0.6282 & Italian & +0.5996 & Thai & +0.5575 \\
7 & Danish & +0.6272 & Asturian & +0.5961 & Italian & +0.5551 \\
8 & Galician & +0.6262 & Vietnamese & +0.5928 & Traditional Chinese & +0.5516 \\
9 & Swedish & +0.6255 & Galician & +0.5904 & Indonesian & +0.5478 \\
10 & Malay & +0.6215 & Indonesian & +0.5841 & Vietnamese & +0.5455 \\
\midrule
\addlinespace[3pt]
\multicolumn{7}{@{}l}{\textbf{(c) Receiving languages, question and answer in the named language}} \\
\addlinespace[2pt]
Rank & Language & Gain & Language & Gain & Language & Gain \\
\midrule
1 & Danish & +0.3044 & Spanish & +0.2652 & English & +0.2189 \\
2 & Vietnamese & +0.3006 & Portuguese & +0.2623 & Spanish & +0.1840 \\
3 & Swedish & +0.2983 & French & +0.2606 & French & +0.1801 \\
4 & Catalan & +0.2976 & Italian & +0.2577 & Italian & +0.1761 \\
5 & Romanian & +0.2950 & English & +0.2463 & Portuguese & +0.1684 \\
6 & Italian & +0.2936 & German & +0.2426 & German & +0.1659 \\
7 & Spanish & +0.2923 & Vietnamese & +0.2342 & Dutch & +0.1359 \\
8 & French & +0.2903 & Indonesian & +0.2321 & Vietnamese & +0.1349 \\
9 & Dutch & +0.2875 & Thai & +0.2192 & Romanian & +0.1213 \\
10 & Galician & +0.2844 & Romanian & +0.2188 & Norwegian Bokm{\aa}l & +0.1182 \\
\bottomrule
\end{tabular*}
\endgroup
\end{table}

\FloatBarrier

\subsection{Unlearning transfer and remaining-access bottlenecks}
\label{app:findings-unlearning-transfer}
\begin{table}[htbp]
\centering
\caption{\textbf{A few languages are the recurring bottlenecks of unlearning transfer.} For each acquisition-language unlearned checkpoint, the other language with the largest positive remaining acquired-access fraction $R=(p_{\mathrm{U}}-p_{\mathrm{O}})/(p_{\mathrm{FT}}-p_{\mathrm{O}})$, with question and answer both in that language, is its bottleneck. Entries count how often each language is the bottleneck over the 70 checkpoints (23 Aya, 21 Qwen, 26 Llama; per-model counts in parentheses), top five per method with ties. ``None'' counts checkpoints in which every eligible alternative is at or below the oracle.}
\label{tab:destination-language-bottlenecks-compact}
\label{tab:destination-language-bottlenecks}
\begingroup\small
\setlength{\tabcolsep}{4pt}\renewcommand{\arraystretch}{1.1}
\begin{tabularx}{\textwidth}{@{}lYr@{}}
\toprule
Method & Most frequent bottleneck languages: checkpoints /70 (Aya/Qwen/Llama) & None \\
\midrule
SimNPO   & Kannada 5 (0/0/5), North Levantine Arabic 4 (1/2/1), Kamba 4 (2/0/2), Galician 3 (0/1/2), Gujarati 3 (3/0/0), Russian 3 (2/1/0), Simplified Chinese 3 (0/2/1) & 0 \\
\addlinespace[2pt]
GradDiff & North Levantine Arabic 4 (1/3/0), Russian 4 (1/1/2), Simplified Chinese 4 (0/3/1), Wolof 3 (3/0/0), Ta'izzi-Adeni Arabic 2 (1/0/1), Moroccan Arabic 2 (0/0/2), Crimean Tatar 2 (2/0/0), English 2 (0/0/2), Galician 2 (0/1/1), Occitan 2 (1/1/0) & 12 (2/1/9) \\
\addlinespace[2pt]
RMU      & Russian 27 (8/4/15), English 10 (0/2/8), Arabic 6 (6/0/0), Ta'izzi-Adeni Arabic 4 (2/2/0), Simplified Chinese 4 (0/4/0) & 0 \\
\bottomrule
\end{tabularx}
\par\vspace{3pt}
\begin{minipage}{\textwidth}\footnotesize
Eligible routes require $p_{\mathrm{FT}}\geq0.10$ and $p_{\mathrm{FT}}-p_{\mathrm{O}}\geq0.05$; Aya's Armenian parent has no eligible alternative. Counts are pooled checkpoint counts, not equal-family percentages or generated-answer disclosure rates.
\end{minipage}
\endgroup
\end{table}

\FloatBarrier

\subsection{Agreement between acquisition and unlearning transfer}
\label{app:findings-transfer-agreement}
\begin{table}[htbp]
\centering
\caption{\textbf{Languages that gain the most access during learning also lose the most during unlearning.} Median within-checkpoint Spearman rank correlation between fine-tuning gain over base $p_{\mathrm{FT}}-p_{\mathrm{B}}$ and the subsequent correct-answer likelihood decrease $p_{\mathrm{FT}}-p_{\mathrm{U}}$ after unlearning in the same language, across 21 training-language checkpoints per model and the two answer-language settings shown.}
\label{tab:acquisition-removal-rank-agreement}
\begingroup\small
\setlength{\tabcolsep}{7pt}\renewcommand{\arraystretch}{1.13}
\begin{tabular*}{\textwidth}{@{\extracolsep{\fill}}lrrr@{}}
\toprule
Unlearning method & Aya & Qwen & Llama \\
\midrule
\multicolumn{4}{@{}l@{}}{\textit{Question translated; answer stays in the training language}} \\
\addlinespace[2pt]
SimNPO & 0.99 & 0.98 & 0.99 \\
GradDiff & 0.99 & 0.97 & 0.98 \\
RMU & 0.96 & 0.86 & 0.92 \\
\addlinespace[6pt]
\multicolumn{4}{@{}l@{}}{\textit{Question and answer translated}} \\
\addlinespace[2pt]
SimNPO & 0.86 & 0.76 & 0.71 \\
GradDiff & 0.90 & 0.77 & 0.60 \\
RMU & 0.74 & 0.45 & 0.35 \\
\bottomrule
\end{tabular*}
\par\vspace{3pt}
\begin{minipage}{\textwidth}\footnotesize
Questions use an alternative language in both panels. Each correlation compares
eligible destinations within one checkpoint; entries are medians across 21
training languages. Eligibility requires fine-tuned likelihood at least 0.10
and excess over the retain-only reference at least 0.05: all 173 alternatives
in the first panel and 36--117 in the second, depending on the checkpoint.
Values near 1 indicate similar language rankings, not percentages removed.
Both differences contain the fine-tuned score; agreement does not establish
more complete forgetting.
\end{minipage}
\endgroup
\end{table}

\FloatBarrier

\subsection{Canonical, paraphrased, and translated questions}
\label{app:findings-expression}
\begin{table}[htbp]
\centering
\caption{\textbf{Modal-verb, subordination and sentence-modality paraphrases leave the most access after unlearning.} English paraphrase types ranked by matched-question likelihood win rate, averaged over competing types and model families. Before/After denote acquisition/post-SimNPO; family ranks refer to After (1 = highest post-unlearning likelihood win rate).}
\label{tab:expression-paraphrase-type-ranking}
\begingroup
\small
\setlength{\tabcolsep}{2pt}
\renewcommand{\arraystretch}{1.12}
\begin{tabular*}{\textwidth}{@{\extracolsep{\fill}}rlrrrrrr@{}}
\toprule
 & & & \multicolumn{2}{c}{Mean win rate (\%)} & \multicolumn{3}{c}{Rank after SimNPO} \\
\cmidrule(lr){4-5}\cmidrule(lr){6-8}
Rank & Paraphrase type & QA & Before & After & Aya & Qwen & Llama \\
\midrule
1 & Modal verb & 28 & 77.7 & 67.6 & 1 & 1 & 1 \\
2 & Subordination & 63 & 54.5 & 58.8 & 3 & 3 & 2 \\
3 & Sentence modality & 100 & 53.7 & 58.1 & 2 & 2 & 3 \\
4 & Same-polarity habitual & 91 & 52.1 & 47.4 & 4 & 5 & 6 \\
5 & Same-polarity contextual & 92 & 53.0 & 45.7 & 8 & 7 & 4 \\
6 & Syntax/discourse structure & 89 & 37.3 & 44.2 & 5 & 4 & 8 \\
7 & Synthetic/analytic & 61 & 45.3 & 43.4 & 7 & 9 & 5 \\
8 & Voice (diathesis) & 26 & 38.8 & 43.0 & 6 & 6 & 9 \\
9 & Word order & 80 & 37.6 & 41.7 & 9 & 8 & 7 \\
\midrule
-- & Format & 6 & -- & -- & -- & -- & -- \\
\bottomrule
\end{tabular*}
\par\vspace{3pt}
\begin{minipage}{\textwidth}\footnotesize
A win means higher correct-answer token likelihood on a shared question; ties count $1/2$. QA is total type coverage; each pair shares 12--92 QA bundles. Format lacks complete pairwise coverage and is unranked.
\end{minipage}
\endgroup
\end{table}

\FloatBarrier

\subsection{Hindi question form and answer-language policy details}
\label{app:findings-form-policy}
\begin{table}[htbp]
\centering
\small
\caption{Question form changes which access routes are suppressed.
The matched-form advantage is the mean additional answer-NLL increase when
training and evaluation both use native Hindi questions or both use Romanized
Hindi, compared with crossing the two forms. The contrast is in answer-NLL units; positive values indicate greater suppression with matched forms. Answers remain in native Hindi.
The 16 runs cross two model families, two seeds, two methods and two training
forms at a fixed 100-update budget. Intervals bootstrap the 20 authors in the
318-fact common assessment panel.}
\label{tab:hindi-script-intervention}
\begin{tabular}{lcc}
\toprule
Method & Matched-form advantage (NLL) $\uparrow$ & 95\% interval \\
\midrule
SimNPO & 6.05 & [5.83, 6.27] \\
RMU & 8.26 & [7.66, 8.80] \\
\bottomrule
\end{tabular}
\end{table}

\FloatBarrier

\clearpage
\section{Source selection: full grids and supporting analyses}
\label{app:source-selection}

Supporting material for \autoref{sec:source-selection}. Residual $R$ is defined in \autoref{sec:protocol}; the primary score averages native routes of the held-out language set. Retain damage is $p_{\mathrm{FT}}-p_{\mathrm{U}}$ averaged over the six native retain routes on 50 retain facts per language.

\subsection{All-language forget supervision}
\label{app:all-language-supervision}

Supervising every acquired language is the naive fix to cross-lingual leakage.
We ran it on all nine development parents under the same stopping rule and
protocol as the $K\leq3$ grids, and saved the selected endpoint. All nine met
the rule, at 75--125 updates; none was censored at the cap.
Table~\ref{tab:all-language-supervision-budget} compares it with English-only
supervision and with the retain-constrained best triple.

\begin{table}[htbp]
\centering
\caption{\textbf{All-language forget supervision removes best but degrades every
language's generations.} Forget supervision in $K$ of the six acquired languages,
averaged over the nine development parents. Supervising one language leaves the
fact recoverable in the others; supervising all six removes it, with more than
five times the degenerate output.}
\label{tab:all-language-supervision-budget}
\begingroup
\small
\setlength{\tabcolsep}{5pt}
\renewcommand{\arraystretch}{1.15}
\begin{tabular*}{\textwidth}{@{\extracolsep{\fill}}lrrr@{}}
\toprule
 & \multicolumn{2}{c}{Forget generations} & \\
\cmidrule(lr){2-3}
Forget supervision & Full leakage $\downarrow$ & Gibberish $\downarrow$ & Retain damage $\downarrow$ \\
\midrule
English only ($K=1$)   & 0.553 & 0.031 & 0.129 \\
Best triple ($K=3$)    & 0.196 & 0.129 & 0.105 \\
All six ($K=6$)        & 0.137 & 0.169 & 0.130 \\
\bottomrule
\end{tabular*}
\endgroup
\end{table}

\subsection{Headroom at every budget}
\label{app:headroom-all-budgets}

\begin{table}[htbp]
\centering
\scriptsize
\setlength{\tabcolsep}{3.5pt}
\caption{All 27 family/inventory/budget grids, seed 1. Uniform is the mean over all subsets at that budget. Residual columns report unscaled $R$ (lower is better). Relative reduction is $100(R_{\mathrm{Uniform}}-R_{\mathrm{best}})/R_{\mathrm{Uniform}}$, in percent (higher is better). Gap CI low is the lower endpoint of the pointwise 95\% paired-author bootstrap interval for $R_{\mathrm{Uniform}}-R_{\mathrm{best}}$, in unscaled $R$ units, with the empirical best subset held fixed. Bootstrap win rate is the percentage of 2,000 paired-author draws in which that subset remains first.}
\label{tab:source-headroom-full}
\begin{adjustbox}{max width=\textwidth}
\begin{tabular}{llcllcccccc}
\toprule
Inventory & Family & $K$ & Best subset & Worst subset & Best $R$ & Uniform $R$ & Worst $R$ & Relative reduction (\%) & Gap CI low & Bootstrap win rate (\%) \\
\midrule
Script-diverse & Aya & 1 & EN & JA & 0.718 & 0.793 & 0.909 & 9.4\% & 0.048 & 87\% \\
 &  & 2 & JA+RU & JA+ZH & 0.492 & 0.570 & 0.705 & 13.6\% & 0.033 & 43\% \\
 &  & 3 & AR+JA+ES & EN+ES+ZH & 0.285 & 0.383 & 0.505 & 25.5\% & 0.073 & 24\% \\
\addlinespace[2pt]
 & Qwen & 1 & EN & JA & 0.715 & 0.815 & 0.939 & 12.3\% & 0.076 & 100\% \\
 &  & 2 & ES+ZH & AR+JA & 0.469 & 0.589 & 0.730 & 20.4\% & 0.092 & 66\% \\
 &  & 3 & AR+JA+ES & EN+ES+ZH & 0.262 & 0.366 & 0.549 & 28.5\% & 0.066 & 31\% \\
\addlinespace[2pt]
 & Llama & 1 & EN & JA & 0.773 & 0.864 & 0.918 & 10.6\% & 0.067 & 100\% \\
 &  & 2 & EN+RU & AR+JA & 0.541 & 0.701 & 0.811 & 22.8\% & 0.124 & 99\% \\
 &  & 3 & EN+AR+JA & EN+RU+ES & 0.410 & 0.516 & 0.643 & 20.5\% & 0.061 & 50\% \\
\addlinespace
Latin-script & Aya & 1 & ID & TR & 0.441 & 0.719 & 0.871 & 38.7\% & 0.227 & 100\% \\
 &  & 2 & ID+SW & DE+TR & 0.258 & 0.398 & 0.561 & 35.1\% & 0.094 & 88\% \\
 &  & 3 & EN+DE+TR & EN+ID+VI & 0.114 & 0.174 & 0.288 & 34.3\% & 0.009 & 49\% \\
\addlinespace[2pt]
 & Qwen & 1 & EN & TR & 0.642 & 0.778 & 0.909 & 17.5\% & 0.103 & 100\% \\
 &  & 2 & DE+ID & ID+VI & 0.367 & 0.479 & 0.609 & 23.4\% & 0.067 & 50\% \\
 &  & 3 & EN+SW+TR & ID+SW+VI & 0.209 & 0.305 & 0.410 & 31.4\% & 0.060 & 64\% \\
\addlinespace[2pt]
 & Llama & 1 & EN & TR & 0.746 & 0.845 & 0.903 & 11.8\% & 0.073 & 100\% \\
 &  & 2 & EN+DE & SW+TR & 0.396 & 0.530 & 0.760 & 25.3\% & 0.065 & 84\% \\
 &  & 3 & EN+DE+VI & EN+ID+SW & 0.093 & 0.282 & 0.515 & 67.0\% & 0.141 & 95\% \\
\addlinespace
Low-resource & Aya & 1 & EN & NE & 0.746 & 0.840 & 0.918 & 11.2\% & 0.069 & 100\% \\
 &  & 2 & AR+NE & HI+NE & 0.500 & 0.600 & 0.765 & 16.7\% & 0.064 & 84\% \\
 &  & 3 & EN+AR+NE & EN+HA+SW & 0.256 & 0.387 & 0.613 & 33.8\% & 0.100 & 39\% \\
\addlinespace[2pt]
 & Qwen & 1 & EN & HA & 0.760 & 0.871 & 0.934 & 12.7\% & 0.079 & 100\% \\
 &  & 2 & EN+HA & HI+NE & 0.584 & 0.728 & 0.906 & 19.8\% & 0.097 & 93\% \\
 &  & 3 & AR+NE+SW & HI+NE+SW & 0.081 & 0.363 & 0.715 & 77.5\% & 0.217 & 65\% \\
\addlinespace[2pt]
 & Llama & 1 & EN & NE & 0.737 & 0.875 & 0.923 & 15.8\% & 0.079 & 100\% \\
 &  & 2 & EN+HI & HI+NE & 0.545 & 0.672 & 0.932 & 18.9\% & 0.086 & 58\% \\
 &  & 3 & EN+HA+HI & AR+HI+NE & 0.309 & 0.467 & 0.746 & 33.8\% & 0.111 & 29\% \\
\bottomrule
\end{tabular}
\end{adjustbox}
\end{table}

\subsection{Cross-request ranking stability}
\label{app:source-request-stability}

\begin{table}[htbp]
\centering
\caption{\textbf{Source-set rankings recur across training orders and across disjoint forget requests} (TOFU, low-resource inventory, $K{=}3$). Cross-request $\rho$ is the Spearman correlation between the rankings of two disjoint forget requests after averaging each subset over its five orders; within-request $\rho$ is the median of the ten pairwise correlations between order-specific rankings; shared top five counts subsets in both requests' top five. Four-to-one selects the lowest-residual subset on four orders of the same request and evaluates it on the held-out fifth (gain in $100R$ over uniform).}
\label{tab:source-request-stability-wide}
\label{tab:source-request-stability}
\begingroup
\small
\setlength{\tabcolsep}{4pt}
\renewcommand{\arraystretch}{1.1}
\begin{tabular*}{\textwidth}{@{\extracolsep{\fill}}lcccrc@{}}
\toprule
 & \multicolumn{3}{c}{Ranking agreement} & \multicolumn{2}{c}{Leave-one-order-out} \\
\cmidrule(lr){2-4}\cmidrule(lr){5-6}
Model / request & \makecell{Cross-request\\$\rho$} & \makecell{Within-request\\$\rho$} & \makecell{Shared\\top five} & \makecell{Four-to-one gain\\mean [min, max]} & \makecell{Positive\\held-out orders} \\
\midrule
Qwen / A     & 0.929 & 0.802 & 3/5 & $+7.11$ [$+0.51$, $+11.97$]  & 5/5 \\
Qwen / B     & 0.668 & 0.760 & 1/5 & $+5.72$ [$+2.69$, $+10.75$]  & 5/5 \\
Tiny Aya / A & 0.923 & 0.840 & 4/5 & $+10.55$ [$+5.77$, $+13.36$] & 5/5 \\
Tiny Aya / B & 0.950 & 0.838 & 5/5 & $+9.84$ [$+7.03$, $+12.88$]  & 5/5 \\
Llama / A    & 0.913 & 0.776 & 3/5 & $+9.24$ [$+4.16$, $+13.72$]  & 5/5 \\
Llama / B    & 0.892 & 0.815 & 3/5 & $+9.26$ [$+5.83$, $+12.30$]  & 5/5 \\
\bottomrule
\end{tabular*}
\endgroup
\end{table}

\subsection{Generation behavior of the selected source sets}
\label{app:selector-behavior}

The behavioral scores in \autoref{tab:selector-original-behavior} are computed on free-form generations rather than on likelihoods. Generations are produced for the native routes of the six inventory languages (query language equal to the requested answer language) at the selected endpoint of each run, for the forget facts of 20 authors and for 50 retain facts per language, and scores are averaged over three training orders; retain scores use six languages for Aya and five (excluding Arabic) for Qwen and Llama. Generations are graded by the judge described in \autoref{app:generation-judge}, which also defines full leakage, any disclosure, retained recovery and extraction strength (ES). ES is reported descriptively, without confidence intervals.

\paragraph{Individual top-3 protocol.}
This comparison uses TOFU and three paired training orders. Individual top-3 selects EN/RU/ES for Aya and EN/ES/ZH for Qwen and Llama; \method selects JA/ES/ZH for Aya and EN/AR/JA for Qwen and Llama. Unlearning uses SimNPO, learning rate $2\times10^{-5}$. Relative to Individual top-3, lower full leakage and higher retained recovery hold in all three orders for Llama and two of three for Aya; Qwen's full-leakage differences change sign across orders.

\subsection{Optional source weighting}
\label{app:source-weighting}

\paragraph{Source set weighting.} Here we weight the forget loss update according to cross-language gradient alignment. We measure gradients with the benign calibration data. Define $g_{i,\ell,b}$ as the gradient of the mean answer-token NLL for sample $i$ in language $\ell$ with respect to decoder block $b$, with $B_\theta$ decoder blocks. We define
\[
C_{s,t}
=\frac{1}{n_G B_\theta}
\sum_{i=1}^{n_G}\sum_{b=1}^{B_\theta}
\max\!\left(0,\cos(g_{i,s,b},g_{i,t,b})\right).
\]
This measures shared loss sensitivity across languages, which we use as a proxy for transfer of forgetting updates. We then choose source weights to maximize the worst held out language's weighted coverage:
\[
\pi^\star
\in\operatorname*{arg\,max}_{\pi\in\mathcal W}
\min_{t\in T(\widehat S)}
\sum_{s\in\widehat S}\pi_s C_{s,t}.
\]
Here $\mathcal W$ contains source weights in $[0.20,0.45]$ that sum to 1. We scale each forget example's SimNPO loss by $K\pi_s^\star$, where $s$ is its language.

\paragraph{Experimental setup.}
We compare Qwen and Llama on TOFU on  three paired training orders, using the same source stopping criterion in \autoref{sec:protocol}. English-heavy weights are $(.45,.275,.275)$ for English, Arabic and Japanese. Our permutation control swaps the calibrated AR/JA weights.

\begin{table}[!htbp]
\centering
\caption{\textbf{Source-weighting comparisons.} Rates (\%) averaged over three orders on TOFU.  Retain uses all six languages in (a) and the indicated held-out languages in (b)}
\label{tab:source-weighting-behavior}
\begingroup\small\setlength{\tabcolsep}{4pt}
\textbf{(a) All six native routes}\par\smallskip
\begin{tabular*}{\linewidth}{@{\extracolsep{\fill}}lrrrrr@{}}
\toprule
Allocation & Full $\downarrow$ & Any $\downarrow$ & Retain $\uparrow$ & Gibberish $\downarrow$ & Target-only $\uparrow$ \\
\midrule
\multicolumn{6}{@{}l}{\emph{Qwen: EN/AR/JA}} \\
Uniform & 34.33 & 57.02 & 82.78 & 2.26 & 87.94 \\
Calibrated & 29.61 & 52.07 & 83.78 & 3.91 & 90.31 \\
English-heavy & 31.67 & 53.94 & 80.67 & 3.11 & 88.24 \\
AR/JA-permuted & 30.94 & 52.74 & 85.00 & 2.17 & 86.72 \\
\multicolumn{6}{@{}l}{\emph{Llama: EN/AR/JA}} \\
Uniform & 31.76 & 57.13 & 85.78 & 5.69 & 83.09 \\
Calibrated & 26.33 & 47.65 & 87.33 & 18.48 & 67.31 \\
\multicolumn{6}{@{}l}{\emph{Qwen: EN/AR/ZH}} \\
Uniform & 34.44 & 60.09 & 80.56 & 0.63 & 90.63 \\
Calibrated & 25.65 & 51.65 & 81.44 & 0.43 & 91.91 \\
\bottomrule\end{tabular*}\par\medskip
\textbf{(b) Own omitted native routes}\par\smallskip
\begin{tabular*}{\linewidth}{@{\extracolsep{\fill}}lrrrrr@{}}
\toprule
Allocation & Full $\downarrow$ & Any $\downarrow$ & Retain $\uparrow$ & Gibberish $\downarrow$ & Target-only $\uparrow$ \\
\midrule
\multicolumn{6}{@{}l}{\emph{Qwen: EN/AR/JA; targets RU/ES/ZH}} \\
Uniform & 41.33 & 62.85 & 86.89 & 1.04 & 93.63 \\
Calibrated & 35.70 & 58.59 & 86.89 & 1.89 & 94.15 \\
English-heavy & 38.85 & 59.96 & 84.89 & 1.41 & 93.48 \\
AR/JA-permuted & 36.70 & 58.30 & 88.22 & 1.26 & 92.81 \\
\multicolumn{6}{@{}l}{\emph{Llama: EN/AR/JA; targets RU/ES/ZH}} \\
Uniform & 42.59 & 67.74 & 90.00 & 0.48 & 91.78 \\
Calibrated & 37.00 & 61.44 & 90.44 & 6.44 & 81.33 \\
\multicolumn{6}{@{}l}{\emph{Qwen: EN/AR/ZH; targets JA/RU/ES}} \\
Uniform & 39.00 & 63.63 & 81.78 & 0.07 & 93.89 \\
Calibrated & 29.59 & 56.63 & 81.11 & 0.04 & 93.19 \\
\bottomrule\end{tabular*}\par\medskip
\endgroup\end{table}
\FloatBarrier

\subsection{Source-set rankings}
\label{app:source-rankings}

\begin{figure}[tbp]
\centering
\begin{minipage}{0.48\textwidth}
  \centering
  \includegraphics[width=\textwidth]{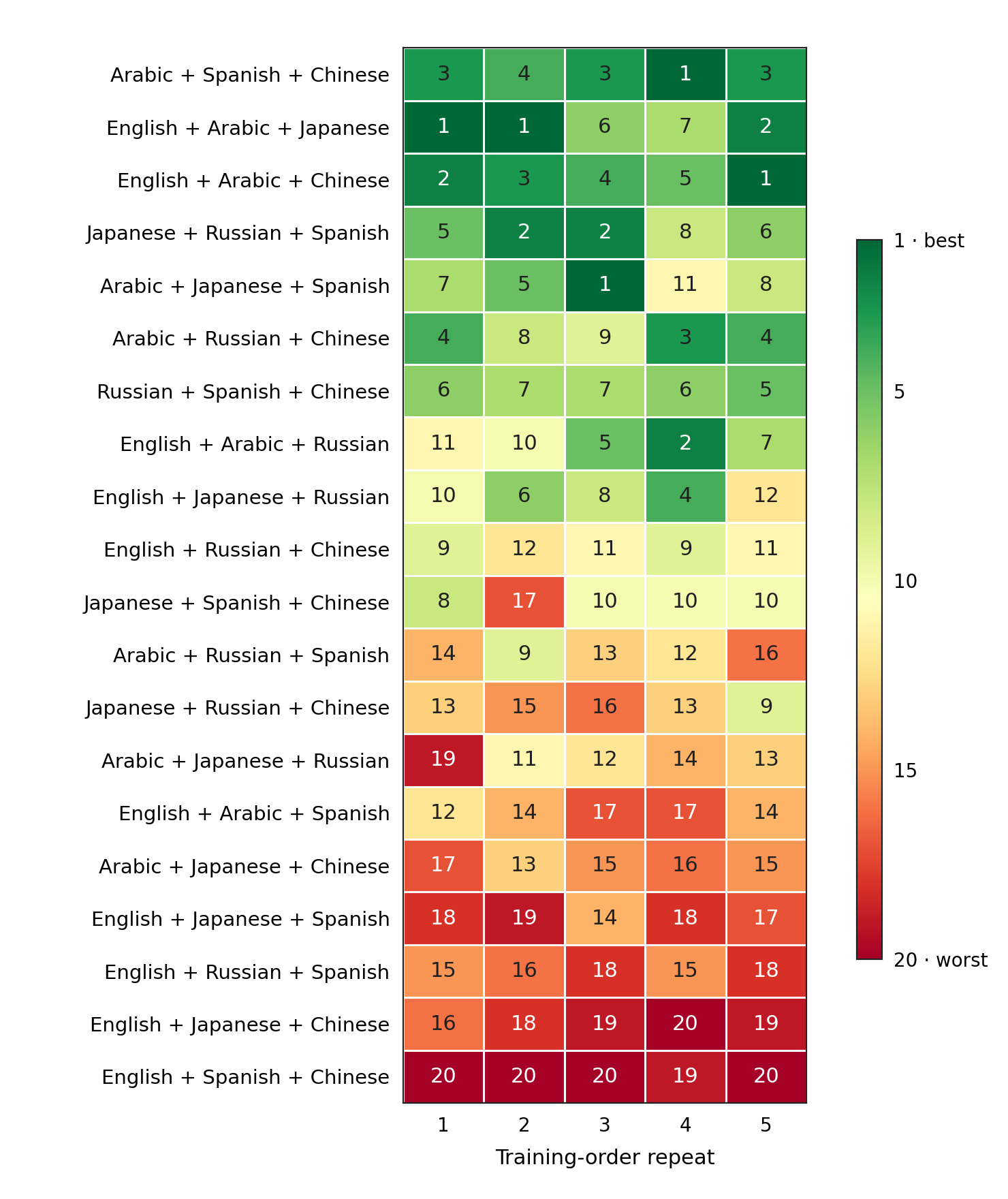}\\[2pt]
  {\small (a) Qwen, script-diverse inventory}
\end{minipage}\hfill
\begin{minipage}{0.48\textwidth}
  \centering
  \includegraphics[width=\textwidth]{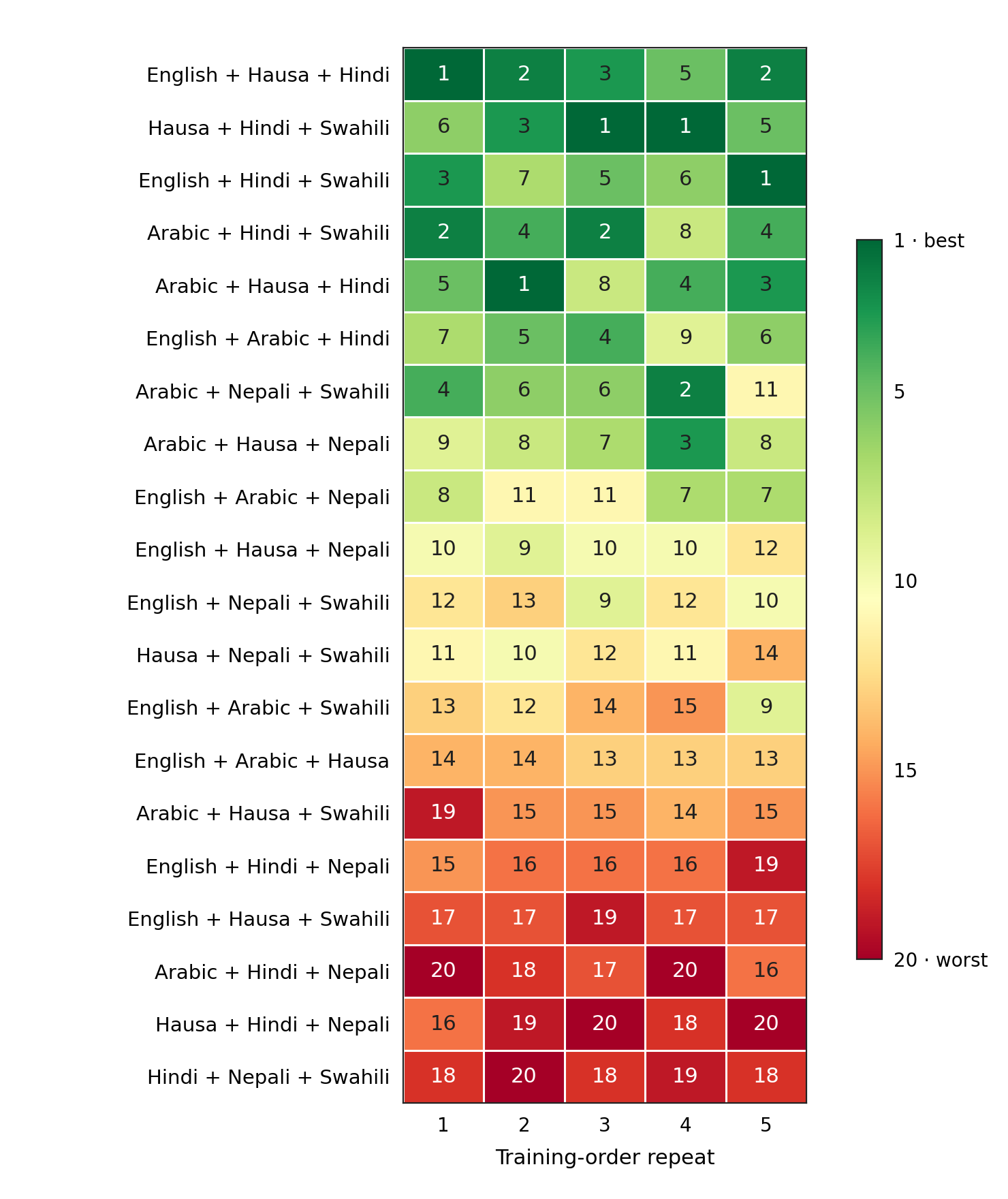}\\[2pt]
  {\small (b) Qwen, low-resource inventory}
\end{minipage}

\vspace{8pt}

\begin{minipage}{0.48\textwidth}
  \centering
  \includegraphics[width=\textwidth]{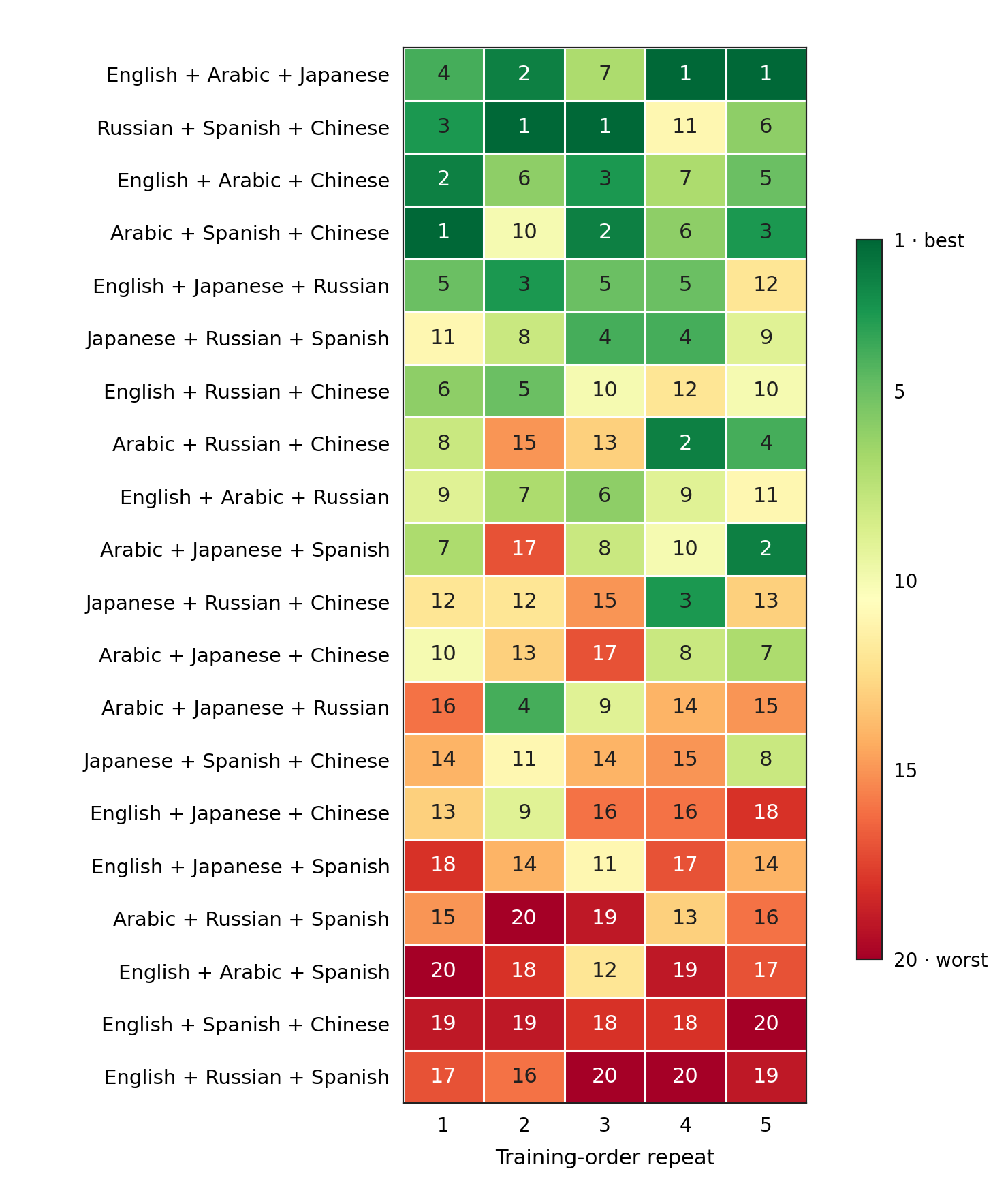}\\[2pt]
  {\small (c) Llama, script-diverse inventory}
\end{minipage}\hfill
\begin{minipage}{0.48\textwidth}
  \centering
  \includegraphics[width=\textwidth]{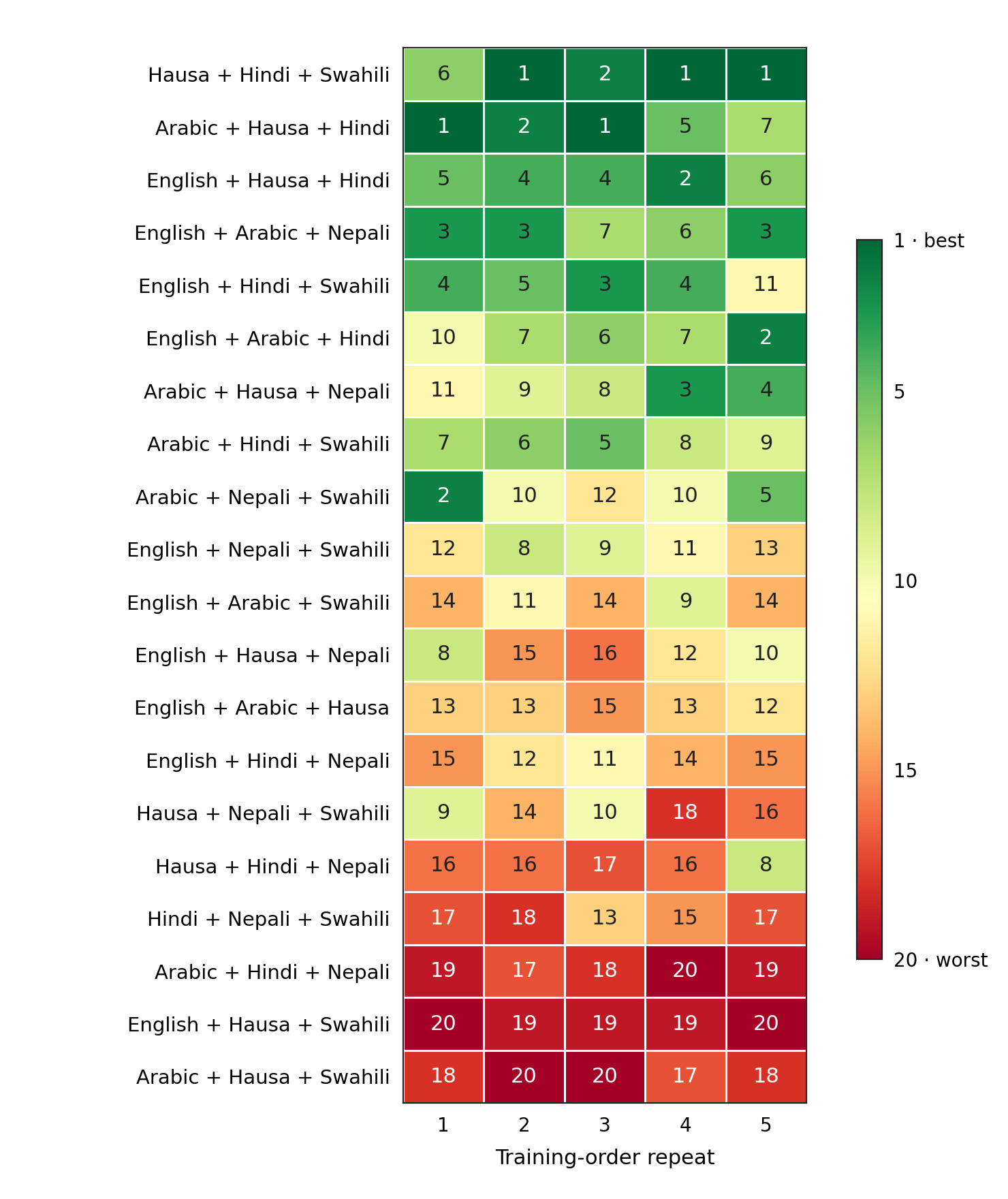}\\[2pt]
  {\small (d) Llama, low-resource inventory}
\end{minipage}

\caption{\textbf{Good and bad source sets recur across training orders.} Each panel ranks the 20 three-language subsets (rows, ordered by their mean rank) by held-out residual access under each of five training orders (columns); 1 is best and 20 worst. The script-diverse inventory is English, Arabic, Japanese, Chinese, Russian and Spanish; the low-resource inventory is English, Arabic, Hausa, Hindi, Nepali and Swahili. The top and bottom of each ranking recur across orders while the middle is less stable, and the low-resource inventory (b, d) is the more stable of the two, in line with the consistency values of \autoref{tab:development-ranking-consistency}.}
\label{fig:findings-rank-heatmaps}
\end{figure}

\FloatBarrier
\clearpage
\section{Reproducibility}
\label{app:reproducibility}

\begin{itemize}
    \item \textbf{Training.}
    \begin{itemize}
        \item \textbf{Models and acquisition.}
        We use Tiny Aya Global, Qwen3-4B-Instruct-2507, and
        Llama-3.2-3B-Instruct. TOFU fine-tuning uses learning rate $2\times10^{-5}$, effective batch 64, and sequence length 512, with five epochs for multilingual checkpoints and six for translated monolingual checkpoints.
        \item \textbf{Unlearning settings.}  We use paged AdamW with 32-bit optimizer states, weight decay $0.01$, effective batch 16, and constant learning rates after warmup.
        \item \textbf{TOFU monolingual benchmark.} Training uses sequence length 512, 20 warmup updates, and a 400-update cap. SimNPO and GradDiff use learning rate $2\times10^{-5}$; RMU uses $10^{-5}$. SimNPO uses $\beta=3.5$, $\delta=1$, forget-loss coefficient $0.25$, and retain-NLL coefficient $1$. GradDiff uses forget/retain coefficients $1/5$. RMU targets decoder block 7 (zero-indexed), with steering norm 2 and equal forget/retain coefficients.
        \item \textbf{TOFU source selection and weighting.} Both use the same SimNPO loss settings above, sequence length 768, 50 warmup updates, and a 600-update cap. Learning rates are $2\times10^{-5}$ for source selection and $10^{-5}$ for weighting. Training uses FP32 parameters with BF16 autocast.
    \end{itemize}
    \item \textbf{Endpoint selection.}
    \begin{itemize}
        \item Stopping uses 100 forget facts per language;
    \end{itemize}
    \item \textbf{Calibration}
    \begin{itemize}
        \item \method uses two disjoint panels of 100 retained examples each for TOFU and 60 for LUME.
    \end{itemize}
\end{itemize}

\end{document}